\documentclass[journal,twoside,web]{ieeecolor}
\usepackage{generic}
\usepackage{cite}
\pdfoutput=1
\usepackage{amsmath,amssymb,amsfonts}
\usepackage{algorithmic}
\usepackage{graphicx}
\usepackage{textcomp}
\def\BibTeX{{\rm B\kern-.05em{\sc i\kern-.025em b}\kern-.08em
    T\kern-.1667em\lower.7ex\hbox{E}\kern-.125emX}}
\usepackage{bm}
\usepackage{algorithm}
\usepackage{stfloats}
\usepackage{subfig}
\usepackage{autobreak}
\usepackage{graphicx}
\usepackage{multirow}

\usepackage{soul}

\graphicspath{ {./Fig/} }
\usepackage{textcomp}

\usepackage[colorlinks,
linkcolor=blue,
anchorcolor=blue,
citecolor=blue]{hyperref}
\usepackage[all]{hypcap} 
\def\BibTeX{{\rm B\kern-.05em{\sc i\kern-.025em b}\kern-.08em
    T\kern-.1667em\lower.7ex\hbox{E}\kern-.125emX}}

\begin{document}
\title{Hybrid Control Policy for Artificial Pancreas via Ensemble Deep Reinforcement Learning}

\author{Wenzhou Lv, Tianyu Wu, Luolin Xiong, Liang Wu, Jian Zhou, \\ Yang Tang \IEEEmembership{Senior Member, IEEE}, and Feng Qian
\thanks{Wenzhou Lv, Tianyu Wu, Luolin Xiong, Yang Tang, and Feng Qian are with the Key Laboratory of Smart Manufacturing in Energy Chemical Process, Ministry of Education, East China University of Science
and Technology, Shanghai 200237, China (e-mail: lyuwenzhou@gmail.com; tianyuwu813@gmail.com; xiongluolin@gmail.com; 
tangtany@gmail.com; fqian@ecust.edu.cn).}
\thanks{Liang Wu and Jian Zhou are with the Metabolism, Shanghai Jiao Tong University Affiliated Sixth People’s Hospital, Shanghai Diabetes Institute, Shanghai Clinical Center for Diabetes, Shanghai
200233, China (e-mail: bill\_wuliang@hotmail.com; zhoujian@sjtu.edu.cn).}
}

\maketitle

\begin{abstract}
\textit{Objective}: The artificial pancreas (AP) has shown promising potential in achieving closed-loop glucose control for individuals with type 1 diabetes mellitus (T1DM). However, designing an effective control policy for the AP remains challenging due to the complex physiological processes, delayed insulin response, and inaccurate glucose measurements. While model predictive control (MPC) offers safety and stability through the dynamic model and safety constraints, it lacks individualization and is adversely affected by unannounced meals. Conversely, deep reinforcement learning (DRL) provides personalized and adaptive strategies but faces challenges with distribution shifts and substantial data requirements. 
\textit{Methods}: We propose a hybrid control policy for the artificial pancreas (HyCPAP) to address the above challenges. HyCPAP combines an MPC policy with an ensemble DRL policy, leveraging the strengths of both policies while compensating for their respective limitations. To facilitate faster deployment of AP systems in real-world settings, we further incorporate meta-learning techniques into HyCPAP, leveraging previous experience and patient-shared knowledge to enable fast adaptation to new patients with limited available data.
\textit{Results}: 
We conduct extensive experiments using the FDA-accepted UVA/Padova T1DM simulator across three scenarios. Our approaches achieve the highest percentage of time spent in the desired euglycemic range and the lowest occurrences of hypoglycemia.
\textit{Conclusion}: The results clearly demonstrate the superiority of our methods for closed-loop glucose management in individuals with T1DM.
\textit{Significance}: The study presents novel control policies for AP systems, affirming the great potential of proposed methods for efficient closed-loop glucose control.

\end{abstract}



\section{Introduction}
\label{sec:introduction}
    \IEEEPARstart{T}{ype} 1 diabetes mellitus (T1DM) is a chronic autoimmune disease characterized by insulin deficiency resulting from the destruction of pancreatic $\beta$-cells, leading to persistent high blood glucose (BG) levels known as hyperglycemia \cite{1}. 
    Therefore, individuals with T1DM require the administration of exogenous insulin to achieve better glucose regulation. To assist in managing glucose levels, the artificial pancreas (AP) has been developed. The AP is a closed-loop system that integrates a continuous glucose monitoring (CGM) device and an insulin pump, enabling automated insulin delivery based on real-time BG measurements \cite{2}. The primary objective of the AP is to maintain safe and healthy BG levels through closed-loop control algorithms, thereby preventing episodes of hypoglycemia and hyperglycemia in individuals with T1DM.
    \begin{figure}
        \centering
        \includegraphics[width=0.485 \textwidth ]{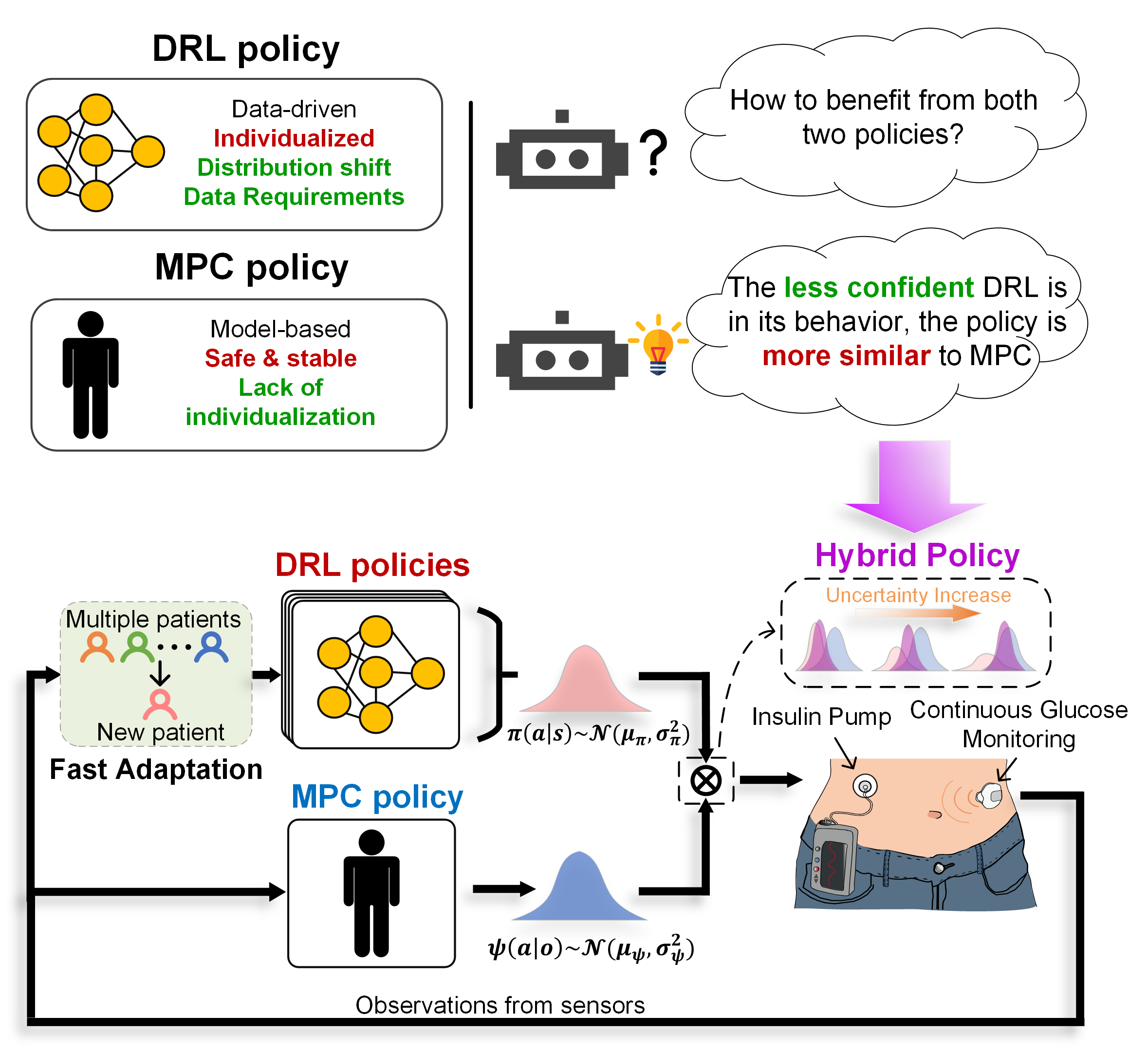}
        \caption{Motivation and overview of the proposed method. The inclusion of the fast adaptation is optional and contingent upon the available amount of data.}
        \label{fig:motivation}
    \end{figure}
    
    \par However, regulating BG levels in real-time is challenging due to the complex physiological processes, delayed insulin response, and inaccurate CGM measurements. To address these challenges, various control algorithms have been developed, such as the proportional-integral-derivative (PID) \cite{PID-1, PID-2, PID-3, PID-4}, model predictive control (MPC) \cite{MPC-1, MPC-2, magni-RW, MPC-4-RW, MPC-5-RW, MPC-6-RW, zone_MPC}, and fuzzy logic (FL) \cite{FL-1,FL-2,FL-3}. Of these approaches, PID and MPC are the most widely used and have been leveraged for commercial AP systems in real life \cite{Commercial-AP}. PID is relatively simple and easy to implement for glucose control, but it may face limitations in handling nonlinear dynamics and responding effectively to meal disturbances \cite{PID-Short}. MPC is another control approach that employs a mathematical model of the specific-patient dynamics to predict future glucose levels and minimizes a given cost function by solving a constrained optimization problem. 
    
    \par Compared to PID, MPC has shown greater potential in dealing with nonlinear dynamics and managing glucose levels, as demonstrated by numerous clinical studies \cite{MPC-PID}. By incorporating system dynamics and safety constraints, MPC provides secure and stable insulin delivery. However, the performance of MPC heavily relies on the accuracy of the personalized dynamic model, which necessitates invasive and expensive parameter tuning \cite{tune-MPC_1, tune-MPC_2}. Despite efforts to individualize the patient model using available clinical information, accurately capturing an individual's glucose-insulin dynamics remains a significant challenge. This patient-model mismatch can lead to limited effectiveness in glucose control. Additionally, unannounced meals, where no meal bolus insulin injection is administered, can adversely affect the prediction accuracy of the dynamic models, resulting in undesirable glucose control outcomes. Therefore, addressing these challenges is crucial for improving the performance and reliability of MPC-based AP policies.

    \par Reinforcement learning (RL) is a direct approach for adaptive optimal control of nonlinear systems \cite{RL}. Recently, deep reinforcement learning (DRL) \cite{DRL}, which incorporates deep learning and RL, has emerged as a powerful technique for addressing sequential decision-making tasks \cite{decision_1,decision_2,decision_3}. Numerous studies have demonstrated that DRL is a promising method for developing AP systems compared to traditional approaches \cite{RL-1,RL-2,SAC_DDPG,RL-4-RW,RL-5-RW,RL-6-RW,Fox-RW}. However, it is important to note that the DRL may encounter the distribution shift problem, resulting in unpredictable decisions when faced with unfamiliar or previously unseen states. In safety-critical glucose management problems, such unpredictable decisions can potentially lead to fatal failures for patients. Therefore, addressing the distribution shift problem becomes crucial to ensure the reliability and safety of DRL-based AP systems. Furthermore, training a DRL policy typically requires a substantial amount of data, which can pose challenges in real-world settings where data collection is often limited. 

    \par We observe that certain advantages of MPC and DRL are complementary. While MPC ensures safety and stability, DRL enhances individualization and adaptability as depicted in Fig. \ref{fig:motivation}. By combining the strengths of both approaches, we can potentially overcome their individual limitations and achieve a more comprehensive and robust AP system. Motivated by the limitations of existing control policies and the above discoveries, our aim is to develop a control policy that combines the benefits of both policies and extends it to adapt to new patients with limited data. To this end, we present HyCPAP, a hybrid control policy for the AP system. HyCPAP integrates a prior MPC policy and an ensemble DRL policy, which leverages multiple DRL policies generated through Masksembles\cite{Masksembles}, an uncertainty estimation technique. Combining the two policies in an unsuitable manner not only fails to leverage the strengths of each policy but also undermines their overall effectiveness, consequently leading to suboptimal glycemic decisions. Therefore, we combine the two policies utilizing the uncertainty of policy. When the multiple DRL policies exhibit different behaviors, HyCPAP performs similarly to the MPC policy. This characteristic allows HyCPAP to mitigate distribution shifts and ensure secure and reliable performance, particularly in unfamiliar scenarios. On the other hand, when the multiple DRL policies exhibit similar behaviors, HyCPAP leverages the strengths of the DRL policy, complementing the lack of individualization observed in the MPC policy. By combining the strengths of both policies, HyCPAP compensates for their individual shortcomings, leading to more secure and individualized glucose management. Additionally, in order to facilitate faster deployment of AP systems in real-world settings and address the data requirements of DRL, we further incorporate meta-learning techniques into HyCPAP, leveraging previous experience and patient-shared knowledge. This enables our policy to fast adapt to new patients with limited data. We refer to the case where sufficient training data is available as \textbf{the general case} and where specific patient's data is limited as \textbf{the data-limited case}. We refer to the method for the data-limited case as Meta-HyCPAP. The overview of our proposed method is illustrated in Fig. \ref{fig:motivation}.  

    Specifically, the main contributions of our work can be summarized as follows:
        \begin{itemize}
            \item We propose a hybrid control policy for the AP system, named HyCPAP. HyCPAP combines a prior MPC policy with an ensemble DRL policy. By integrating the strengths of both policies, HyCPAP provides personalized and adaptive capabilities while concurrently ensuring safety and stability.
            \item We further present Meta-HyCPAP, an extension of HyCPAP that integrates meta-learning techniques. By leveraging previous experience and shared knowledge among patients, Meta-HyCPAP enables fast adaptation to new patients with limited available data.
            \item We conduct comprehensive experiments using the FDA-accepted UVA/Padova T1DM simulator \cite{Simulator}. Our proposed methods are compared against the Zone-MPC [2], Soft Actor Critic (SAC) [3], and Meta-SAC baseline methods. The results clearly demonstrate the effectiveness of our methods in glucose management, as they achieve the highest percentage of time spent in the desired euglycemic range, and the lowest occurrences of hypoglycemia, in both general and data-limited cases. These findings indicate the great potential of our methods for closed-loop glucose control in individuals with T1DM.
        \end{itemize}

    \par The remainder of this paper is organized as follows: Section \ref{sec:related_work} introduces the related work for glucose control. Section \ref{sec:methodology} provides a detailed formulation of the glucose control problem and presents the proposed HyCPAP and Meta-HyCPAP methods. Section \ref{sec:experiments} provides an overview of the experimental settings. Section \ref{sec:results} presents the experimental results, accompanied by comprehensive discussions. Finally, Section \ref{sec:conclusion} concludes the paper.


\section{Related Work}
    \label{sec:related_work}
    \begin{figure*}[!t]
        \centering
        \includegraphics[width=0.91 \textwidth ]{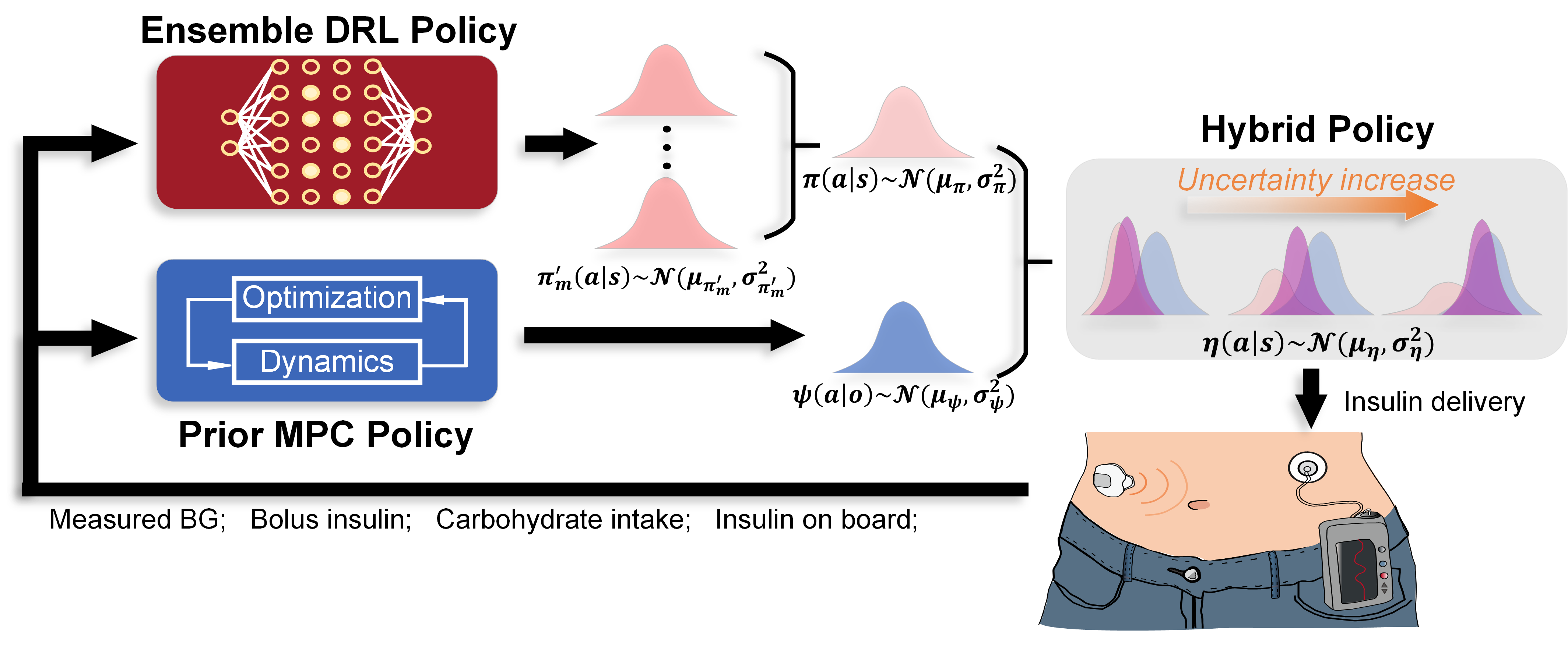}
        \caption{Framework of our hybrid policy. 1) Ensemble DRL policy $\pi{(a \vert s)}$ is a mixture of a set of SAC policies $\pi^{\prime}(a \vert s)$ via Masksembles, an uncertainty estimation technique. 2) The hybrid policy combines the prior MPC policy and the ensemble DRL policy. As the level of uncertainty increases, the hybrid policy progressively aligns its behavior with that of the MPC policy.} 
        \label{fig:method_overall}
    \end{figure*}
    In this section, we review the MPC-based and DRL-based glucose control policies. Subsequently, we will discuss the limitations of existing research in this field. Additionally, we will provide an overview of our proposed method.
    \subsection{MPC-based glucose control policy} 
        A variety of MPC policies have been proposed in previous studies for closed-loop glucose regulation. In \cite{magni-RW}, an average patient model is utilized to design and validate a linear MPC control system with the aim of maintaining glucose levels at a specific set point. 
        In order to further improve the accuracy of the patient model, individualization patient models have been designed. In \cite{MPC-4-RW}, a control-relevant model is developed that incorporates individual patient information and effectively captures an individual's response to insulin. To enhance the effectiveness of MPC policies, researchers have explored various modifications to the cost function of MPC. Recognizing that euglycemia is defined as a range, an innovative Zone-MPC control policy is proposed in \cite{MPC-5-RW}, which sets the control target as the glucose zone instead of a set-point. Moreover, some studies have explored the inclusion of terms related to glucose velocity into the cost function of MPC \cite{MPC-6-RW,zone_MPC}. 
        
        MPC ensures secure and stable insulin delivery by incorporating system dynamics and safety constraints. However, its performance heavily relies on the accuracy of the personalized dynamic model. Despite efforts to capture individual glucose-insulin dynamics using clinical information, challenges persist, leading to limited glucose control effectiveness. Unannounced meals further impact dynamic model prediction accuracy, resulting in sub-optimal glucose control outcomes.
    \subsection{DRL-based glucose control policy}
        Data-driven DRL control policies have been introduced to address the challenges of glucose regulation and provide personalized treatment for individuals with T1DM. In \cite{RL-4-RW}, an adaptive control policy based on actor-critic learning is proposed, providing daily updates for the basal rate and insulin-to-carbohydrate ratio. Alternatively, some studies focus on insulin delivery decisions in real-time rather than daily updates. In \cite{RL-5-RW}, a control policy that utilizes the Deep Q-Network (DQN) algorithm for personalized discrete insulin delivery is presented. Additionally, in \cite{RL-6-RW}, a dual-hormone control policy which is also based on DQN that shows improvements compared to a single-hormone control policy. Furthermore, it is suggested by \cite{Fox-RW} to leverage recurrent neural networks for capturing the feature of historical glucose and providing continuous insulin actions.
        
        DRL offers individualized and adaptive control strategies by training policy on specific patient data. However, it is susceptible to the distribution shift problem, whereby the distribution of the training data differs from that encountered during testing. The problem can lead to sub-optimal control performance and potential safety risks, particularly in glucose control problem. Additionally, training a DRL policy typically requires a significant amount of data. However, in real-world scenarios, newly-admitted patients can only provide limited BG data, posing challenges for efficiently training DRL policies to manage glucose levels.
		
        To address these limitations and leverage the complementary advantages of the two policies, we propose HyCPAP, a novel hybrid control policy for the AP system, which combines a prior MPC policy and an ensemble DRL policy. Additionally, to facilitate real-world deployment of AP systems, we further incorporate meta-learning techniques into HyCPAP, enabling fast adaptation for new patients with limited data using previous experience and patient-shared knowledge.


\section{Methodology}
    \label{sec:methodology}
    
    In this section, we formulate the glucose control problem and introduce its key components. Next, we present our proposed method, HyCPAP, which combines a prior MPC policy with a set of DRL policies using an uncertainty estimation technique. Finally, we introduce Meta-HyCPAP, an extension of HyCPAP that incorporates meta-learning for fast adaptation to new patients with limited data. The overall framework of our hybrid policy is depicted in Fig. \ref{fig:method_overall}.
    \subsection{Problem Formulation}
        Considering the unavailability of the patient's true state described by \cite{true-state} and the presence of sensor noise, we frame the glucose control as a partially-observable Markov decision process (POMDP) defined by a tuple $(S, A, \mathcal{T}, R, O, \Omega,\gamma)$, where $S$, $A$, $\mathcal{T}$ and $R$ represent the state space (i.e., physiological state), the action space (i.e., insulin), the transition probability (i.e., physiological model) and the reward function, respectively. In a POMDP, an agent cannot directly obtain the underlying state of environment $s_t$ but instead receives an observation $o_{t+1} \in O$ when transitioning to the next state $s_{t+1}$ with the probability $\Omega(o_{t+1} \vert s_{t+1})$. 
        In this work, we build a recurrent version of SAC \cite{SAC} which takes previous observations, actions, and rewards as additional inputs to handle the POMDP. We leverage the SAC algorithm because of its superior performance for glucose control \cite{SAC_DDPG}. 

        \subsubsection{Observation} 
            The observation $o_t$ at step $t$ consists of BG levels $g_t$, meal bolus insulin $b_t$, carbohydrate meal intake $c_t$, and insulin on board (IOB) $i_t$. The IOB accounts for insulin that has been infused but still working on the body. The $g_t$ is measured by CGM every 5 minutes. And the bolus insulin $b_t$ is calculated from meal intake $c_t$ with a bolus calculator described in \cite{zone_MPC}.

        \subsubsection{Action} 
            The scale of action $a_t$ is subject-specific ranging from 0 to 10 times the basal insulin delivery rate. The basal insulin aims to maintain healthy fasting plasma glucose levels. Additionally, it is easily accessible in clinical settings.

        \subsubsection{Reward Function} 
            Presently, reward functions used in DRL for AP systems are typically based on glucose risk, aiming to maintain BG levels at a steady state. However, according to clinical consensus \cite{glucose-consensus}, the desired BG target should be within the range of [70,180] mg/dL, rather than a fixed value. And the recommended glucose target is within the range of [80,140] mg/dL. Considering the above glucose target and the risk of hypoglycemia, introduce a novel \textbf{zone reward function} to enhance the ability of the trained DRL policy to reduce the time spent in hypoglycemia and improve glucose control efficiency. The zone reward function is expressed as: 
            \begin{equation}
                r_t=\left\{\begin{array}{cc}
                0, & 39\! \leq \! g_{t+1} \! <\! 54 \\
                1-0.201 *\left(100-g_{t+1}\right)^{1.622}, & 54 \! \leq \! g_{t+1} \! < \! 100 \\
                1, & 100 \! \leq \! g_{t+1} \! \leq \! 140 \\
                1-0.473 *\left(g_{t+1}-140\right)^{0.918}, & 140\!<\!g_{t+1} \! \leq \! 300 \\
                0.5. & 300\! < \! g_{t+1} \! \leq \! 400 \\
                \end{array}\right.
            \end{equation}
            where the $g_t$ denotes the BG level in time instant $t$.
            To mitigate the risks of hypoglycemia, we adjust the reward function to prioritize maintaining BG levels between [100, 140] mg/dL, rather than the previous range of [80, 140] mg/dL. Furthermore, BG levels outside the range of [39, 400] mg/dL are considered critical situations that necessitate termination. The penalty for hypoglycemia termination is -20, while for hyperglycemia, it is -10.
        \par The goal of the glucose control problem is to increase the time spent in euglycemia and prevent episodes of hypoglycemia and hyperglycemia in individuals with T1DM.

    
    \subsection{Hybrid Control}
    
        We present a hybrid control policy for AP systems, called HyCPAP, as depicted in Fig. \ref{fig:method_overall}. HyCPAP comprises a model-based Zone-MPC policy and a data-driven ensemble DRL policy which leverages multiple DRL policies. When each DRL policy performs similarly, HyCPAP tends to resemble the DRL policy. Conversely, when the multiple DRL policies exhibit different behaviors, HyCPAP performs similarly to the MPC policy. In this manner, HyCPAP can benefit from both policies and compensate for their shortcomings, providing individualized insulin delivery while ensuring the safety of patients.

        
        \subsubsection{Prior MPC Policy}
            We use the adaptive periodic Zone-MPC with a dynamic cost function described in \cite{zone_MPC} as the prior policy. The policy continuously updates its control penalty parameters in real-time according to the predicted glucose state and its velocity. It operates every 5 minutes driven by CGM. Specifically, at each update time instant $t$, the control law of the Zone-MPC is derived by solving an optimization problem:
            \begin{equation}
                u_{0: N_{u}-1}^*:=\mathop{\arg \min} _{u_{0: N_{u}-1}} J\left(x_t, u_{0: N_{u}-1}\right)
            \end{equation}
            with cost function
            \begin{equation}
                J(\cdot, \cdot):= \! \sum_{k=1}^{N_{y}}\left(\check{z}_k^2\!+\!P\left(v_k\right) \hat{z}_k^2\!+\!\hat{D} \hat{v}_k^2\right) \!+\!\sum_{k=0}^{N_u-1}\left(\hat{R} \hat{u}_k^2\!+\!\check{R}\hat{u}_k^2\right),
            \end{equation}
            and subject to system dynamics and safety constraints. 
            
            The insulin-glucose model state $x_t$ is estimated using the most recent CGM readings. $N_y$ and $N_u$ represent the prediction horizon and the control horizon, respectively. The insulin delivery deviation from the basal rate $u_{\text{basal}}$ is represented by ${u}_k$ and consists of two parts: $\hat{u}_k$ and $\check{u}_k$, which indicate the delivery rate above and below $u_{\text{basal}}$, respectively. The weights $P(v_k)$, $\hat{R}$, and $\Check{R}$ regulate the controller's response to deviations from the target glucose levels, which are represented by $\hat{z_k}$ and $\check{z_k}$. Glucose velocity $v_k$ is linked to the predicted glucose state. The coefficient $\hat{D}$ adjusts the penalty for positive glucose velocity $\hat{v}_{k}$.

            We model the actions of the prior MPC policy as an independent Gaussian distribution ${\psi}(a \vert s) \sim \mathcal{N}(\mu_{\psi},\sigma^2_{\psi})$, where $\mu_{\psi}$ and $\sigma^2_{\psi}$ denote the mean and corresponding variance, respectively. It should be noted that the mean of the distribution equals the absolute insulin delivery rate $\mu_{\psi} = u_k+u_{\text{basal}}$. The variance $\sigma^2_{\psi}$ is determined empirically in this study. For a more comprehensive understanding of the terms and the algorithm, we recommend referring to the work \cite{zone_MPC}.

        \begin{figure*}[]
                \centering
                \includegraphics[width=0.95\textwidth]{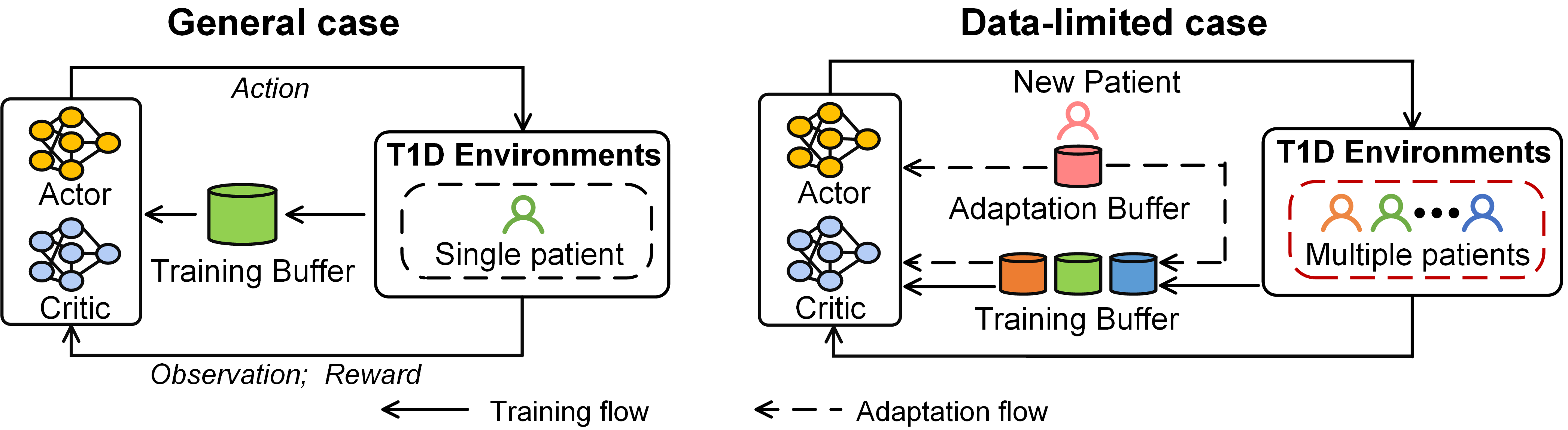}
                \caption{DRL policy generation procedure. The ensemble DRL policy is trained with one specific patient in the general case but is trained with multiple patients in the data-limited case. And for data-limited case, an extra adaptation process that reuses training data is carried out.}
                \label{fig:DRL_generation}
            \end{figure*}
        \subsubsection{Ensemble DRL Policy}
            
            In order to accurately quantify the level of confidence or uncertainty associated with the DRL policy, we incorporate Masksembles \cite{Masksembles} into our neural networks. Masksembles enables uncertainty estimation by generating a limited and fixed set of masks, while also sharing a portion of the network. This approach strikes a balance between Deep Ensemble \cite{deep_ensemble} and Monte-Carlo Dropout \cite{MC_Dropout} techniques. The number and overlapping of these fixed masks regulate the correlation between the submodels. The implementation of Masksembles\cite{Masksembles} is straightforward, as the masks are inserted between two layers in the neural networks.
            
            Then, we leverage SAC \cite{SAC} algorithm, which models each action as an independent Gaussian distribution ${\pi^\prime}(a\vert s) \sim \mathcal{N}(\mu_{\pi^\prime},\sigma^2_{\pi^\prime})$, where $\mu_{\pi^\prime}$ and $\sigma^2_{\pi^\prime}$ denote the mean and variance of the distribution, respectively. By employing Masksembles \cite{Masksembles}, we can generate multiple DRL policies, and combine them to estimate the uncertainty associated with the DRL actions. As seen in Fig. \ref{fig:method_overall}, the mean $\mu_{\pi}$ and variance $\sigma^{2}_{\pi}$ of the ensemble DRL policy, a uniformly weighted Gaussian mixture model, are obtained by combining $M$ individual SAC policies ${\pi^\prime}(a \vert s)$:
                \begin{equation}
                \label{eqn:ensemble_mu}
                    \mu_{\pi}(s) = M^{-1}\sum_{m} \mu_{\pi^{\prime}_{m}}(s),      
                \end{equation}
                
                \begin{equation}
                \label{eqn:ensemble_std}
                    \sigma^{2}_{\pi}=M^{-1} \sum_{m}(\sigma^{2}_{\pi^{\prime}_{m}}(s)+\mu^{2}_{\pi^{\prime}_{m}}(s))-\mu^{2}_{\pi}(s),    
                \end{equation}
            where $\mu_{\pi^{\prime}_{m}}$ and $\sigma^{2}_{\pi^{\prime}_{m}}$ denote the mean and variance of m-th SAC policy ${\pi^{\prime}_{m}}(a \vert s)$. 
            As a result, when the multiple DRL policies behave more differently from one another, the ensemble DRL policy tends to exhibit less confidence, resulting in a greater variance $\sigma^{2}_{\pi}$ in ensemble policy ${\pi}(a\vert s) \sim \mathcal{N}(\mu_{\pi},\sigma^2_{\pi})$. 

        
        \subsubsection{Hybrid Control}
            The hybrid policy HyCPAP inherits the advantages of both policies by utilizing the Gaussian multiplication to combine the ensemble DRL policy and the prior MPC policy. Specifically, HyCPAP follows the composite Gaussian distribution, whose mean and variance  $\eta(a \vert s) \sim \mathcal{N}\left(\mu_\eta, \sigma_\eta^2\right)$ can be expressed as follows:
            
                \begin{equation}
                    \label{eqn:hybrid_1}
                    \mu_\eta=\frac{\mu_\pi \sigma_\psi^2+\mu_\psi \sigma_\pi^2}{\sigma_\psi^2+\sigma_\pi^2},
                \end{equation}
                
                \begin{equation}
                    \label{eqn:hybrid_2}
                    \sigma_\eta^2=\frac{\sigma_\psi^2 \sigma_\pi^2}{\sigma_\psi^2+\sigma_\pi^2} .
                \end{equation}
        
            As indicated in Equations (\ref{eqn:hybrid_1}-\ref{eqn:hybrid_2}), in situations where the standard deviation of actions, denoted by $\sigma_{\pi}$, is high due to unfamiliarity or lack of confidence, HyCPAP $\eta(a \vert s)$ performs similarly to the prior MPC policy ${\psi}(a \vert s)$. This characteristic allows HyCPAP to mitigate distribution shifts and ensure secure and reliable performance. Conversely, when the multiple DRL policies exhibit similar behaviors, HyCPAP leverages the strengths of the DRL policy ${\pi}(a\vert s)$, complementing the lack of individualization observed in MPC. This approach allows HyCPAP to effectively integrate the strengths of both policies and compensate for their respective

    
    \subsection{Meta Adaptation}
    Training a DRL policy for the AP typically requires significant amounts of data. However, data collection for new patients can be costly and time-consuming. Therefore, it is crucial to address data limitations, especially in cases where acquiring extensive patient-specific data may be unfeasible. To this end, we propose Meta-HyCPAP, which inherits the ability of the meta-learning \cite{MQL} to leverage previous experience and task-shared knowledge to enable fast adaptation of the policy to new patients. Meta-HyCPAP combines fine-tuning and training data reuse techniques to further enhance the training of the ensemble DRL policy, allowing it to effectively adapt to new patients with limited data. This approach facilitates faster deployment of AP systems in real-world settings.
    
        
        \subsubsection{Training Procedure} 
            During meta training procedure, we apply a multi-task loss function to train each DRL policy: 
            \begin{equation}
                \label{eqn:meta_train}
                {\theta}_{\text {meta}}=\underset{\theta}{\arg \max } \frac{1}{n} \sum_{k=1}^n \underset{\tau \sim D_k}{\mathbb{E}}\left[\ell^k(\theta)\right],
            \end{equation}
            where $\ell^k(\theta)$ is the objective of the SAC algorithm for the k-th train task.
            The only difference between individual training and meta training is using a set of patients instead of a single patient to train the model as shown in Fig. \ref{fig:DRL_generation}. 
            And the training procedure can be expressed as Algorithm \ref{Alg:Meta_Training}.
            
           \begin{algorithm}
            \caption{Meta Training}
            \label{Alg:Meta_Training}
            \small
            \begin{algorithmic}[1]
                \REQUIRE Set of training tasks $\mathcal{D}_\text{train}$ and learned $M$ SAC policies $\{\pi^{\prime}_{i}(a_i \vert s_i)\ \vert\  i \in \{1,..., M\}\}$ 
                \STATE Initialize the replay buffer $\mathcal{B}$
                \STATE Initialize parameters $\theta$ of the learned policies.
                \FOR{each iteration}
                \STATE Sample a task $\mathcal{D} \sim \mathcal{D}_\text{train}$
                \FOR{each timestep $t$}
                    \STATE Compute the single univariate Gaussian distribution $\pi(a \vert s) \sim \mathcal{N}(\mu_\pi,\sigma^2_{\pi})$ using Equations (\ref{eqn:ensemble_mu}-\ref{eqn:ensemble_std})
                    \STATE Sample action $a_t$ from $\pi(a \vert s)$ and perform it
                    \STATE Store transitions in replay buffer $\mathcal{B}$.
                \ENDFOR
                \FOR{each gradient step}
                    \STATE Sample random mini-batch from buffer $\mathcal{B}$
                    \STATE Update each policy based on SAC algorithm \cite{SAC}
                \ENDFOR
                \ENDFOR
                \STATE $\theta_{meta} \leftarrow \theta$
                \RETURN $\theta_{meta}$, $\mathcal{B}$
            \end{algorithmic}
        \end{algorithm}

        
        \subsubsection{Adaptation Procedure} 
            After meta training, we obtain a set of DRL policies containing task-shared knowledge. The first step is to fine-tune these policies using the limited data from new patients with this function: 
            
            \begin{equation}
                \label{eqn:finetune}
                \underset{\theta}{\arg \max }\left\{\underset{\tau \sim D_\text{new}}{\mathbb{E}}\left[\ell^\text{new}(\theta)\right] \right \},
            \end{equation}
            where $\ell^\text{new}(\theta)$ is the objective of the SAC algorithm on the set of new tasks $\mathcal{D}_{\text{new}}$.
            The second step is to exploit the data from the training buffer. While the meta-training tasks, denoted as $\mathcal{D}_\text{train}$, are disjoint from $\mathcal{D}_\text{new}$, the insulin injection strategy may exhibit similarities across different patients. Therefore, transitions collected during the meta-training procedure can be potentially reused to facilitate the obtained policies to adapt to new patients efficiently.
            
            We fit a logistic classiﬁer for each DRL policy on a mini-batch of transitions from the meta-training replay buffer $\mathcal{B}$ and the transitions collected from the new task. The logistic classifier can estimate the importance ratio $\beta(x)$, which quantifies the odds of a sample $x$ belonging to the training distribution versus the testing distribution \cite{logistic}. This ratio can be used to reweigh data from the meta-training replay buffer for taking updates as
            
            \begin{equation}
            \label{eqn:adaptation}
            \underset{\theta}{\arg \max }\left\{\underset{\tau \sim \mathcal{D}_{\text {new}}}{\mathbb{E}}\left[\beta\left(\tau\right) \ell^{\text {new}}(\theta)\right]-\frac{1-\widehat{\mathrm{ESS}}}{2}\left\|\theta-\widehat{\theta}_{\text {train}}\right\|_2^2\right\}.
            \end{equation}
    
            The normalized Effective Sample Size ($\widehat{\mathrm{ESS}}$) is a measure related to $\beta(x)$ that measures the similarity between the training and testing distributions, described as Equation (\ref{eqn:ESS}). When the two distributions are close to each other, the $\widehat{\mathrm{ESS}}$ approaches one; conversely, when they are far apart, the $\widehat{\mathrm{ESS}}$ approaches zero. As shown in Equation (\ref{eqn:adaptation}), the $\widehat{\mathrm{ESS}}$ is used in the regulation term. The procedure of meta adaptation is expressed in Algorithm \ref{Alg:Meta_Adaptation}.
            \begin{equation}
             \label{eqn:ESS}
               \widehat{\mathrm{ESS}}=\frac{1}{m} \frac{\left(\sum_{k=1}^m \beta\left(x_k\right)\right)^2}{\sum_{k=1}^m \beta\left(x_k\right)^2} \in[0,1].
            \end{equation}

            \begin{algorithm}

                \caption{Meta Adaptation}
                \label{Alg:Meta_Adaptation}
                \small
                \begin{algorithmic}[1]
                    \REQUIRE Test task $\mathcal{D}_{\text{new}}$, training replay buffer $\mathcal{B}$, meta-trained policy $\theta_\text{meta}$
                    \STATE Initialize the temporary buffer $\mathcal{B}_\text{new}$
                    \STATE $\theta \leftarrow \theta_\text{meta}$
                    \STATE $\mathcal{B}_\text{new} \leftarrow$ Gather data from $\mathcal{D}_\text{new}$ using policy $\pi_{\theta_\text{meta}}$
                    \STATE Update parameters $\theta$ using $\mathcal{B}_\text{new}$
                    \STATE Obtain classifier using $\mathcal{B}_\text{new}$ and $\mathcal{B}$.
                    
                    \FOR{each adaptation step}
                        \STATE Sample random mini-batch from training buffer $\mathcal{B}$
                        \STATE Calculate $\beta$ for the mini-batch
                        \STATE Estimate $\widehat{ESS}$ using $\beta$ using Equation (\ref{eqn:ESS}).
                        \STATE Update $\theta$ using Equation (\ref{eqn:adaptation}). 
                    \ENDFOR
                    \STATE Evaluate $\theta$ on a new rollout from test task $\mathcal{D}_\text{new}$
                    \RETURN $\theta$
                \end{algorithmic}
            \end{algorithm}

\section{Experiments}
\label{sec:experiments}
    To demonstrate the effectiveness of our approaches, we conduct experiments on the 10-adult cohort of the FDA-accepted UVA/Padova T1DM simulator \cite{Simulator} across three scenarios. In each scenario, we evaluated a method 20 times for each patient, considering both the announced and unannounced meal situations.

    \subsection{Evaluation Scenarios}
        We design three scenarios to comprehensively evaluate the reward functions and the glucose control policies. The detailed settings for scenarios A-C are as follows:
        \begin{itemize}
            \item \textbf{Scenario A} comprises a 48-hour, six-meal protocol starting at 7:00 on Day 1. Over the two-day period, breakfast (45 g of carbohydrates), lunch (60 g of carbohydrates), and dinner (60 g of carbohydrates) are consumed at 8:00, 13:00, and 19:00, respectively.
            \item \textbf{Scenario B} follows the same meal schedule as Scenario A. However, it should be noted that the carbohydrate consumption in this scenario is significantly higher, and the DRL agent has never encountered this meal size during training. Specifically, breakfast, lunch, and dinner consist of 85g, 100g, and 100g of carbohydrates, respectively. The purpose of Scenario B is to assess the policy's ability to manage extreme meal schedules.
            \item \textbf{Scenario C} aims to evaluate the policy's ability to handle uncertainty protocols related to meal size and meal time. It has a duration of 2 days, with three randomized meals per day of 45 g, 60 g, and 60 g with a coefficient of variation (CV) of 10\%. The meals are scheduled to be consumed at 08:00, 13:00, and 19:00, with a standard deviation (STD) of 60 minutes for each time consumed.
        \end{itemize}

    \subsection{Baselines}
        To evaluate the effectiveness of the proposed reward function, we implement two variants of the SAC algorithm, each employing a specific set-point reward function. These variants employ reward functions derived from the Kovatchev risk \cite{index} and the Magni risk \cite{magni-RW}, respectively. Additionally, we compare our policies against other representative control policies to demonstrate the superiority of our proposed methods for glucose management.
        
            \subsubsection{Reward Function} We utilize two commonly used set-point reward functions as baselines. Despite their different risk calculations, these reward functions share a common expression to calculate the reward, as illustrated below\cite{RL-3}:
            \begin{equation}
                r_t=\left\{\begin{array}{lc}
                \left(-risk\left(g_{t+1}\right)+100\right) / 100, & 39\leq g_{t+1} \geq 400 \\
                -15. & \text else. 
                \end{array}\right.
            \end{equation}
           
            We normalize the reward within the range of [-1, 1] to improve training effectiveness. The termination penalty, as described in \cite{RL-3}, is also scaled accordingly. The risk calculations for the two reward functions are as follows:
            \begin{itemize}
                \item \textbf{Kovatchev} refers to the set-point reward function based on regular glucose risk \cite{index}:
                \begin{equation}
                    risk(g_{t})=10 *\left(1.509 * \ln (g_{t})^{1.084}-5.381\right)^2.
                \end{equation}
                
                \item \textbf{Magni} refers to the set-point reward function based on a more stringent penalty for hypoglycemia glucose risk \cite{magni-RW}:
                \begin{equation}
                    risk(g_t)=10 *\left(3.35506^* \ln (g_t)^{0.8353}-3.7932\right)^2.
                \end{equation}
            \end{itemize}
            
        \subsubsection{Control Algorithm}
            In the general case, we conduct a comparative analysis between HyCPAP, MPC, and SAC. Furthermore, in the data-limited case, we compare Meta-HyCPAP to MPC and Meta-SAC. The specific details of the baseline methods are outlined as follows:
            \begin{itemize}
                \item \textbf{MPC} refers to the adaptive periodic Zone-MPC with a dynamic cost function \cite{zone_MPC}, which has been proven to be both efficient and safe for glucose control. As such, it serves as a robust and reliable baseline for AP control policies.
                \item\textbf{SAC} refers to a recurrent version of SAC \cite{SAC} for continuous action space, incorporating our zone reward function.
                \item\textbf{Meta-SAC} refers to a recurrent version of SAC \cite{SAC} which incorporates our zone reward function and an adaptation process to enable fast adaptation to new patients with limited available data, typically around six days.
            \end{itemize}

    \subsection{Evaluation Metrics}
        We select seven glucose-related metrics to assess the performance of control algorithms. Each of these metrics is described in detail below:
        \begin{itemize}
            \item \textbf{Time in range (TIR)} refers to the percentage of time spent within the clinical glucose target range of [70,180] mg/dL. TIR is widely utilized in clinical practice as an indicator of glucose control effectiveness.
            
            \item \textbf{Time below range (TBR)} refers to the percentage of time spent below a specific glucose level. In our study, we consider two specific thresholds: 70 mg/dL (TBR 1) and 54 mg/dL (TBR 2). TBR 2 is particularly important as it indicates the presence of severe hypoglycemia \cite{severe-Hypoglycemia}. By examining TBR 1 and TBR 2, we can assess the duration and severity of hypoglycemia.
            
            \item \textbf{Time above range (TAR)} refers to the percentage of time spent above a specific glucose level. In this study, we examine two specific glucose levels: 180 mg/dL (TAR 1), and 250mg/dL (TAR 2), indicating severe hyperglycemia \cite{severe-Hyperglycemia}.
            
            \item \textbf{Success rate (SR)} is defined as the percentage of evaluations that result in successful glycemic control. An evaluation is considered unsuccessful if the individual's BG level exceeds 400 mg/dL or falls below 39 mg/dL. By utilizing the SR as a metric, we can effectively assess the safety of the implemented control policies. 
            
            \item \textbf{Control-variability grid analysis (CVGA)} \cite{CVGA} is a metric that evaluates closed-loop control quality in a population by visually representing extreme glucose excursions. Each subject is depicted by a single data point for any given observation period. The CVGA summary outcome provides a numeric assessment of glucose regulation, characterizing overall control in the population. By incorporating CVGA into our analysis, we can comprehensively evaluate and enhance glucose regulation policies for closed-loop control.

        \end{itemize}
        
        The primary objective of achieving safe and effective glucose control is to increase the time spent in euglycemia while avoiding hypoglycemia. Therefore, our evaluation of overall performance across three scenarios focuses on four key metrics: TIR, TBR, SR, and CVGA. We do not consider specific thresholds for TBR or TAR in this analysis. Additionally, we employ all metrics, except for CVGA, for a detailed comparison in each scenario. In this context, we differentiate between two levels of TBR and TAR. We assess normality using the Shapiro-Wilk test and determined statistical significance with either the two-sided paired sample t-test or the two-sided Wilcoxon signed rank test. A p-value lower than 0.05 is considered to indicate statistical significance.
    \begin{figure*}[!htbp]
    \centering
    \subfloat[]{\label{fig:reward_a}\includegraphics[width=0.3 \textwidth ]{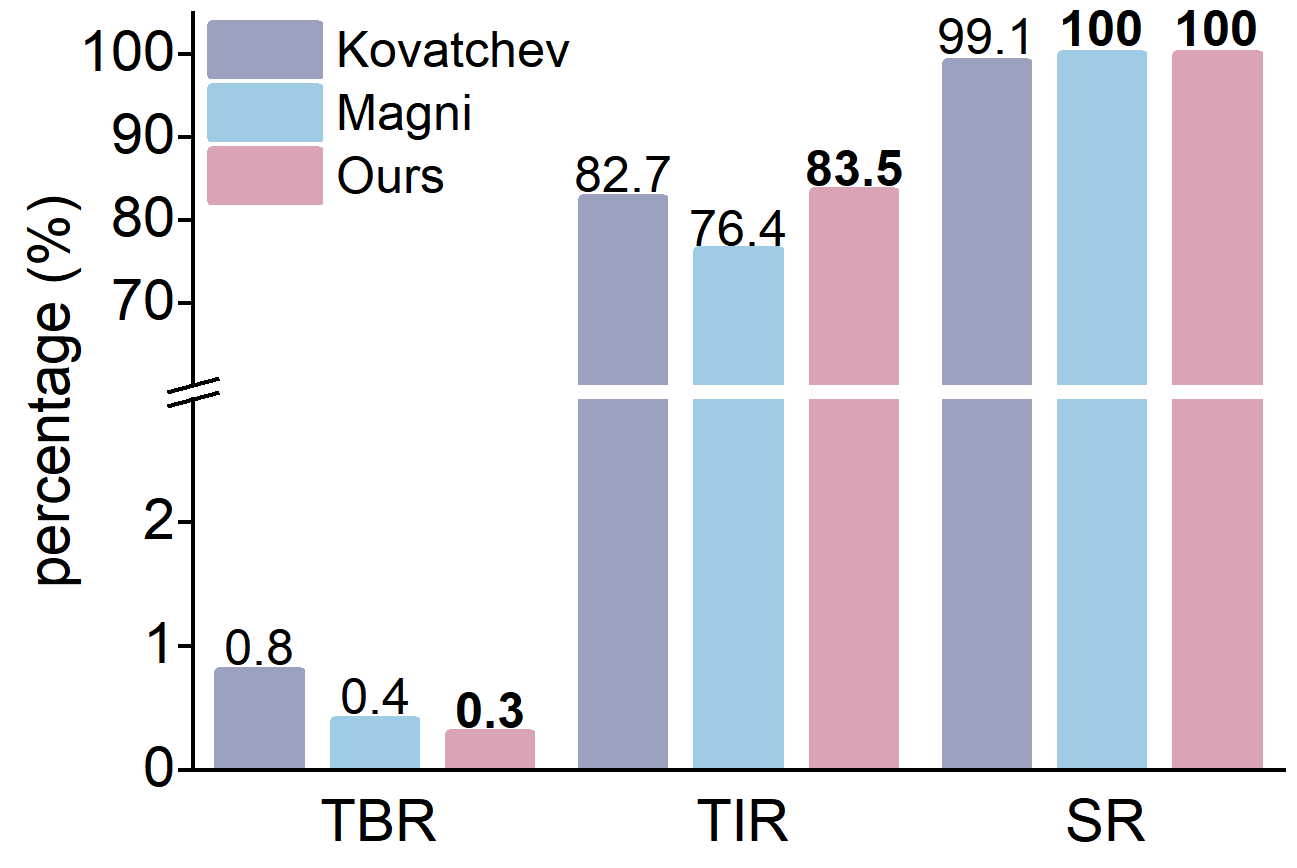}}\hfill
    \subfloat[]{\label{fig:reward_b}\includegraphics[width=0.233 \textwidth, height=0.23\textwidth]{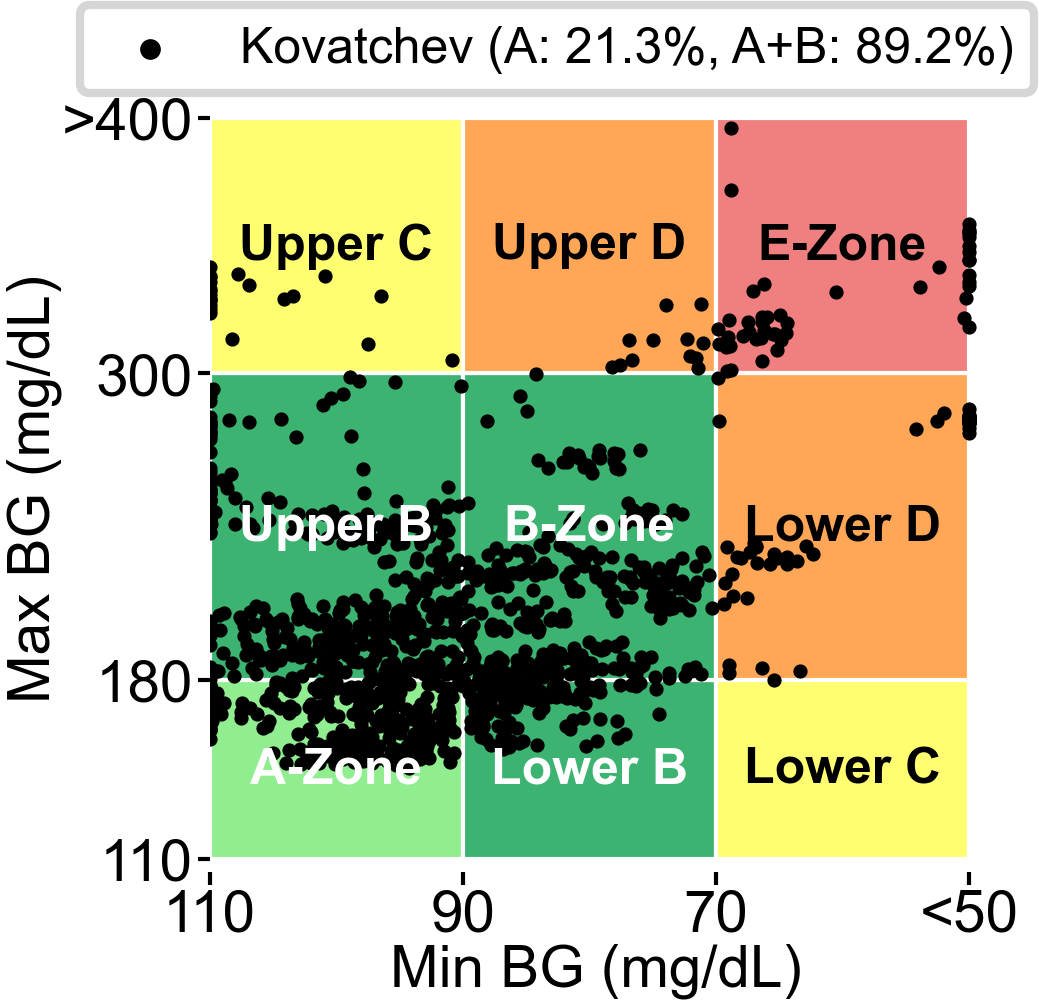}}\hfill
    \subfloat[]{\label{fig:reward_c}\includegraphics[width=0.233 \textwidth, height=0.23\textwidth]{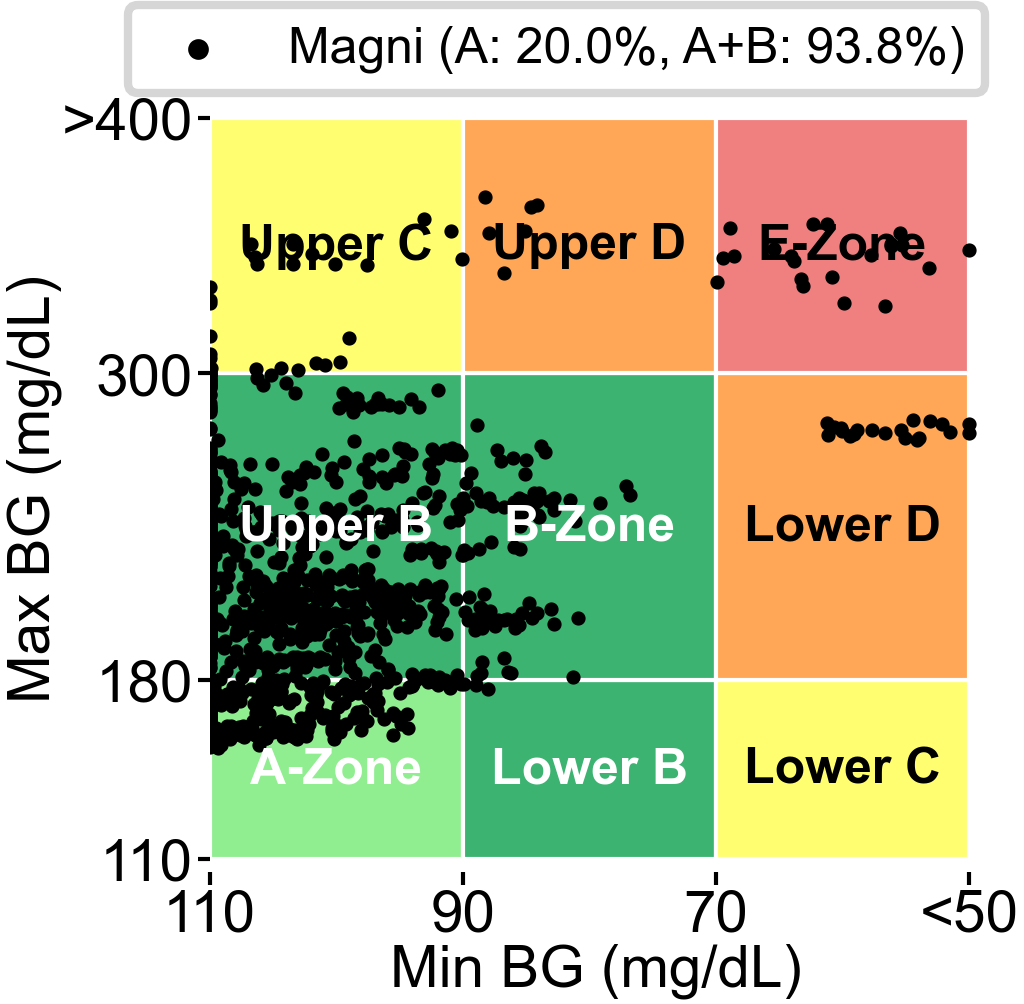}}\hfill
    \subfloat[]{\label{fig:reward_d}\includegraphics[width=0.233 \textwidth, height=0.23\textwidth]{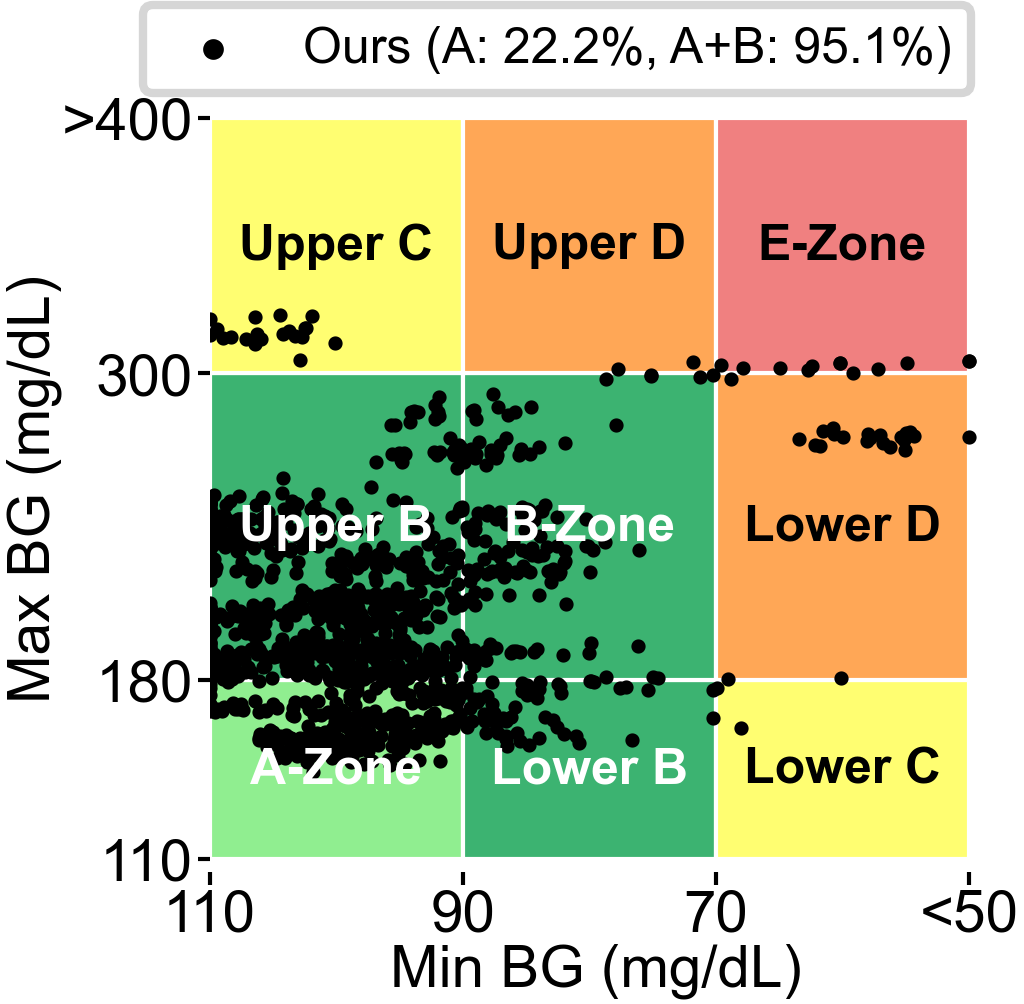}}
    \caption{Overall comparison of reward functions. (a): Time below range (TBR), time in range (TIR), and success rate (SR) of the three reward functions: Kovatchev, Magni, and our proposed zone reward function. The best-performing results are highlighted in bold. (b)-(d): The CVGA plot of Kovatchev, Magni, and our proposed zone reward function, respectively.}
    \label{fig:reward}
    \end{figure*}

\section{Results And Discussion}
    \label{sec:results}
    \subsection{Comparison of Reward Functions}
        In order to demonstrate the superiority of the proposed zone reward function, we conduct a comprehensive comparison between the proposed reward function with two widely used reward functions across three different scenarios. The evaluations of all reward functions are carried out under identical network and hyper-parameter settings.
        
        The overall performance of each reward function is visually depicted in Fig. \ref{fig:reward}. The results clearly indicate that our proposed zone reward function outperforms the other set-point reward functions in various key metrics. It demonstrates the highest percentage of time spent within the target glucose range (\textbf{83.5\%} vs. 82.7\% and 76.4\%). Additionally, it shows a significantly lowest percentage of time spent below the critical threshold of 70 mg/dL (\textbf{0.3\%} vs. 0.8\% and 0.4\%). Moreover, it achieves a remarkable success rate (\textbf{100\%} vs. 99.1\% and \textbf{100\%}), as depicted in Fig. \ref{fig:reward_a}. 
        The superiority of the proposed reward function is further corroborated by Fig. \ref{fig:reward_b}, \ref{fig:reward_c}, and \ref{fig:reward_d}. The proposed reward function achieves the highest percentage in both the A zone (\textbf{22.2\%} vs. 21.3\% and 20.0\%) and the A+B zone (\textbf{95.1\%} vs. 89.2\% and 93.8\%). These results strongly suggest that the proposed zone reward function can enhance the ability of the trained DRL policy to reduce the occurrence of and improve glucose control efficiency.
        
        The zone reward function provides a significant advantage over set-point reward functions due to its ability to penalize deviations of BG predictions from a predefined zone. Unlike set-point reward functions that only consider a single degree of freedom, the zone reward function incorporates two degrees of freedom, represented by upper and lower bounds. This characteristic allows for a range of acceptable glucose levels, taking into account different patient dynamics and individualized treatment goals. The adoption of the zone concept enhances the robustness of DRL policy against plant-model mismatch and errors arising from CGM and provides more personalized and adaptive insulin delivery\cite{zone-concept}. By defining a zone instead of a single set point, the zone reward function acknowledges the natural fluctuations and variability in glucose levels, providing a more personalized and adaptive insulin delivery strategy.
    \subsection{Performance in the General Case}
        In the general case where the training data for each patient is sufficient, our goal is to provide personalized and effective glucose control for each subject. Our extensive results demonstrate that the proposed HyCPAP can leverage the advantages of both MPC and DRL policies while overcoming their respective limitations, achieving the highest time spent with desired glucose levels range against MPC and SAC policies.
        \subsubsection{Scenario Comparison}
\begin{figure}[!htbp]
                \centering
                \includegraphics[scale=0.6]{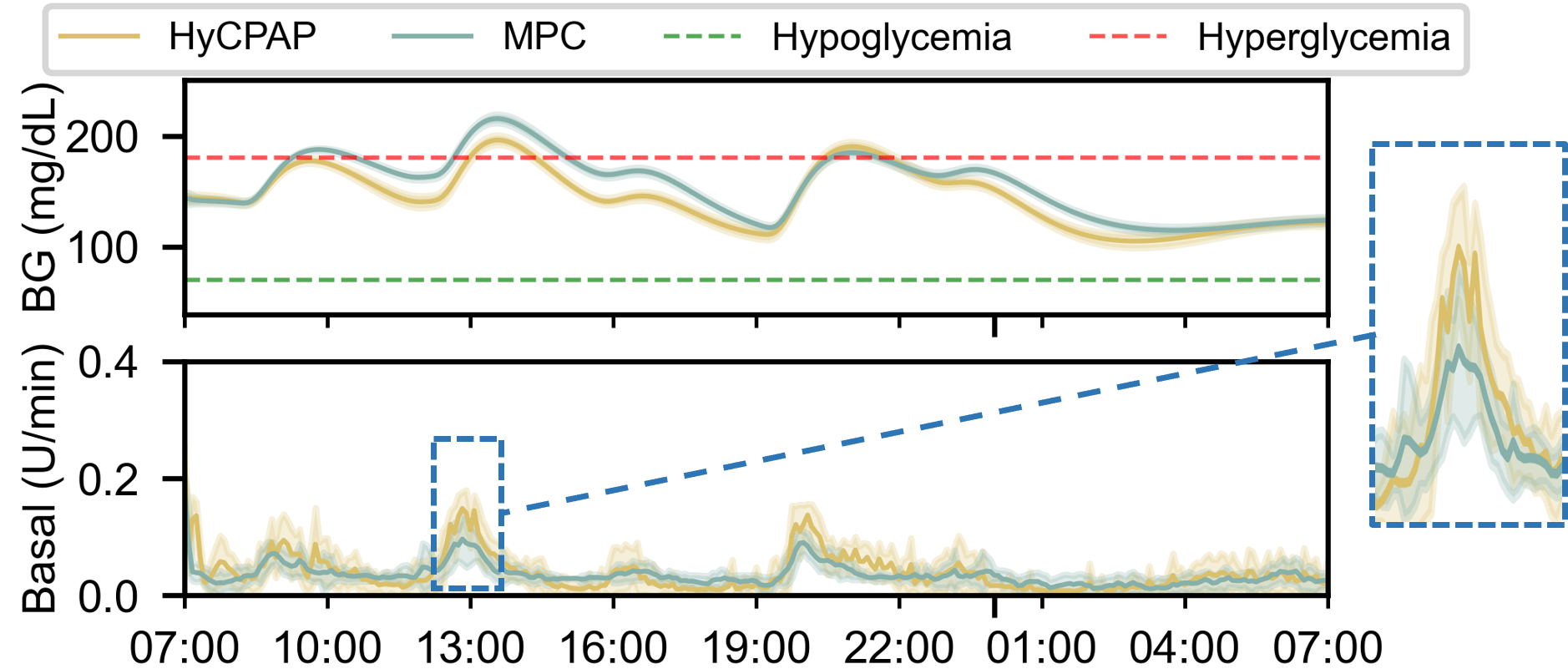}
                \caption{Performance of HyCPAP and MPC on a subject with unannounced meals in scenario A. The meals are consumed at 8:00, 12:00, and 19:00, respectively. The shaded regions indicate the standard deviation. And a magnified view of partial insulin infusion is depicted by the blue dashed box.}
                \label{fig:bg_contrast_UN}
            \end{figure}

\begin{figure}[!htbp]
                \centering
                \includegraphics[scale=0.6]{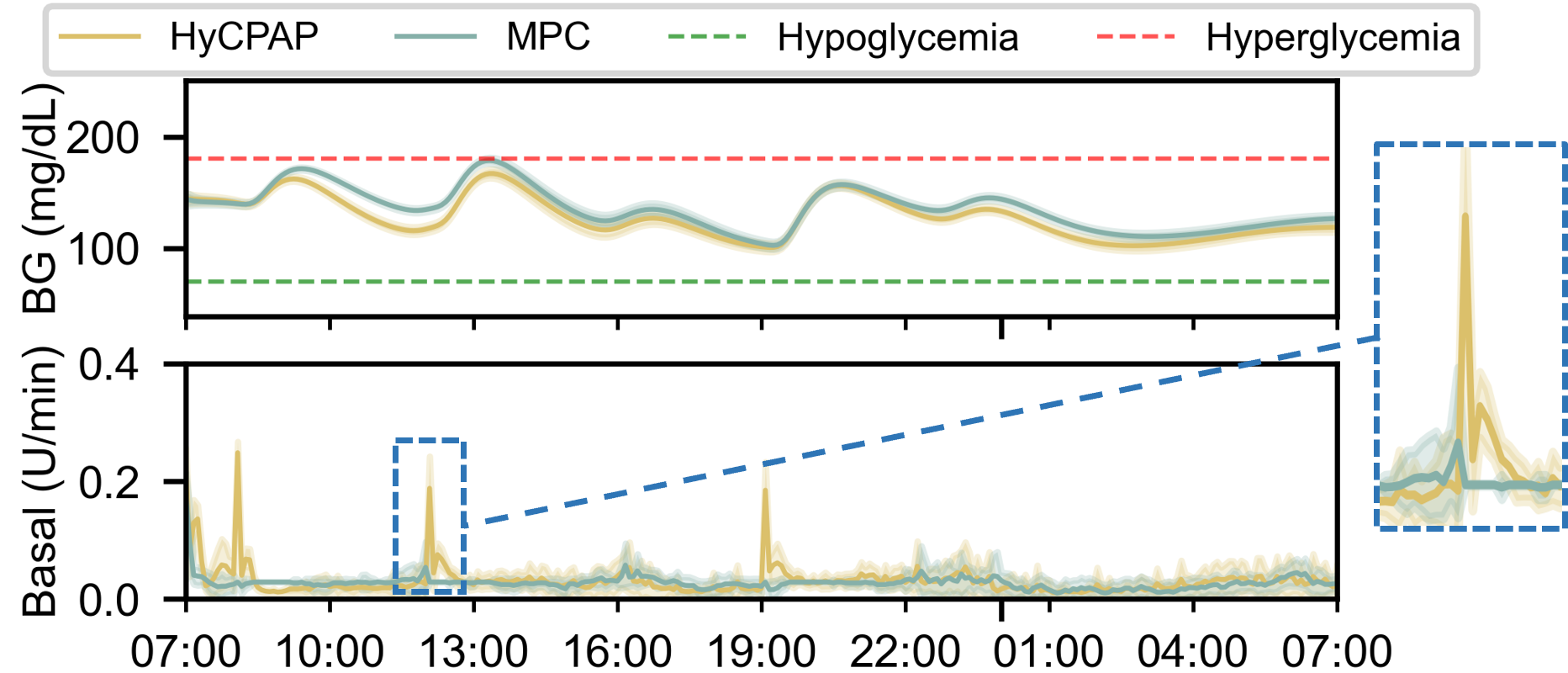}
                \caption{Performance of HyCPAP and MPC on a subject with announced meals in scenario A. The meals are consumed at 8:00, 12:00, and 19:00, respectively. The shaded regions indicate the standard deviation. And a magnified view of partial insulin infusion is depicted by the blue dashed box.}
                \label{fig:bg_contrast}
            \end{figure}

            The results obtained for scenario A are summarized in Table \ref{tab:personalized_a}. While all algorithms showed acceptable performance, HyCPAP consistently outperforms the other two baselines in both announced and unannounced meal situations. Notably, our HyCPAP algorithm achieves superior outcomes in terms of TIR and TAR, indicating its effectiveness in glucose management. The enhanced performance can be attributed to the advantages of the DRL policy. Its inherent flexibility allows for active regulation of insulin delivery in response to rapidly rising glucose levels, particularly in unannounced meal scenarios, as illustrated in Fig. \ref{fig:bg_contrast_UN}. Moreover, the DRL policy can adaptively compensate for calculated bolus insulin delivery, effectively adapting to individual patient requirements and further enhancing control efficacy, as depicted in Fig. \ref{fig:bg_contrast}. Additionally, the utilization of the ensemble technique yields slight improvements compared to using a single SAC policy, as supported by the previous study \cite{sunrise}.
    \begin{table*}[!htbp]
    \centering
    \renewcommand\arraystretch{1.05}
    \caption{Scenario A Metrics in the General Case}
    \label{tab:personalized_a}
    \resizebox{0.97 \textwidth}{!}{%
    \begin{tabular}{lr@{$\pm$}lr@{$\pm$}lr@{$\pm$}l|r@{$\pm$}lr@{$\pm$}lr@{$\pm$}l}
    \hline
    \multicolumn{1}{c}{\multirow{2}{*}{\begin{tabular}[c]{@{}c@{}}Scenario A\\ Metrics\end{tabular}}} &
      \multicolumn{6}{c|}{Announced} &
      \multicolumn{6}{c}{Unannounced} \\ \cline{2-13} 
    \multicolumn{1}{c}{} &
      \multicolumn{2}{c}{MPC} &
      \multicolumn{2}{c}{SAC} &
      \multicolumn{2}{c|}{\textbf{HyCPAP}$\quad$} &
      \multicolumn{2}{c}{MPC} &
      \multicolumn{2}{c}{SAC} &
      \multicolumn{2}{c}{\textbf{HyCPAP}$\quad$} \\ \hline
    TBR 2 (\%) $\downarrow$ &
      0.0 &
      0.0 &
      0.0 &
      0.0 &
      0.0 &
      0.0 &
      0.0 &
      0.0 &
      0.0 &
      0.0 &
      0.0 &
      0.0 \\
    TBR 1 (\%) $\downarrow$ &
      0.0 &
      0.0 &
      0.0 &
      0.2 &
      0.0 &
      0.0 &
      0.0 &
      0.0 &
      0.0 &
      0.0 &
      0.0 &
      0.0 \\
    TIR (\%) $\uparrow$ &
      92.6 &
      9.4 &
      96.9 &
      3.7$^*$$\ $ &
      \textbf{97.2} &
      \textbf{3.9$^{*,\dag}$$\ $} &
      77.4 &
      13.2 &
      81.8 &
      12.3$^*$ &
      \textbf{83.6} &
      \textbf{14.0$^{*,\dag}$} \\
    TAR1 (\%) $\downarrow$ &
      7.4 &
      9.4 $\ $&
      3.1 &
      3.7$^*$ &
      \textbf{2.8} &
      \textbf{3.9$^{*,\dag}$} &
      22.6 &
      13.2 &
      18.2 &
      12.3$^*$ &
      \textbf{16.4} &
      \textbf{14.0$^{*,\dag}$} \\
    TAR2 (\%) $\downarrow$ &
      0.0 &
      0.0 &
      0.0 &
      0.0 &
      0.0 &
      0.0 &
      0.4 &
      0.9 &
      \textbf{0.1} &
      \textbf{0.3$^*$} &
      \textbf{0.0} &
      \textbf{0.2$^*$} \\
    SR (\%) $\uparrow$ &
      \multicolumn{2}{c}{100.0} &
      \multicolumn{2}{c}{100.0} &
      \multicolumn{2}{c|}{100.0$\quad$} &
      \multicolumn{2}{c}{100.0} &
      \multicolumn{2}{c}{100.0} &
      \multicolumn{2}{c}{100.0 $\quad$} \\ \hline
      \multicolumn{13}{l}{\begin{tabular}[c]{@{}l@{}} \scriptsize{Symbols $*$ and $\dag$ indicate statistical significance ($p \leq 0.05$) with respect to the MPC and SAC, respectively. For different} \\ \scriptsize{metrics, $\uparrow$ means the higher the better, and $\downarrow$ contrarily. Results in bold denote the best.}\end{tabular}} 
    \end{tabular}%
    }
    \end{table*}
    
    \begin{table*}[!htbp]
    \centering
    \renewcommand\arraystretch{1.05}
    \caption{Scenario B Metrics in the General Case}
    \label{tab:personalized_b}
    \resizebox{0.97 \textwidth}{!}{%
    \begin{tabular}{lr@{$\pm$}lr@{$\pm$}lr@{$\pm$}l|r@{$\pm$}lr@{$\pm$}lr@{$\pm$}l}
    \hline
    \multicolumn{1}{c}{\multirow{2}{*}{\begin{tabular}[c]{@{}c@{}}Scenario B\\ Metrics\end{tabular}}} &
      \multicolumn{6}{c|}{Announced} &
      \multicolumn{6}{c}{Unannounced} \\ \cline{2-13} 
    \multicolumn{1}{c}{} &
      \multicolumn{2}{c}{MPC} &
      \multicolumn{2}{c}{SAC} &
      \multicolumn{2}{c|}{\textbf{HyCPAP}} &
      \multicolumn{2}{c}{MPC} &
      \multicolumn{2}{c}{SAC} &
      \multicolumn{2}{c}{\textbf{HyCPAP$\quad$}} \\ \hline
    TBR 2 (\%) $\downarrow$ &
      0.0 &
      0.0 &
      0.1 &
      0.1 &
      0.0 &
      0.0 &
      0.0 &
      0.0 &
      0.1 &
      0.1 &
      0.0 &
      0.0 \\
    TBR 1 (\%) $\downarrow$ &
      \textbf{1.1} &
      \textbf{3.4} &
      1.4 &
      3.9$^*$ &
      \textbf{1.0} &
      \textbf{3.0$^{\dag}$} &
      \textbf{0.1} &
      \textbf{0.6} &
      0.5 &
      1.9 &
      \textbf{0.0} &
      \textbf{0.0$^{\dag}$} \\
    TIR (\%) $\uparrow$ &
      80.5 &
      19.8 &
      82.9 &
      18.4$^*$ &
      \textbf{83.7} &
      \textbf{18.7$^{*,\dag}$} &
      57.8 &
      11.8 &
      60.3 &
      11.7 &
      \textbf{62.0} &
      \textbf{13.3$^{*,\dag}$} \\
    TAR1 (\%) $\downarrow$ &
      18.4 &
      17.5 &
      15.7 &
      15.5$^*$ &
      \textbf{15.3} &
      \textbf{16.5$^{*,\dag}$} &
      42.1 &
      11.8 &
      39.3 &
      10.6 &
      \textbf{38.0} &
      \textbf{13.3$^{*,\dag}$} \\
    TAR2 (\%) $\downarrow$ &
      2.0 &
      5.3 &
      0.8 &
      2.2$^*$ &
      \textbf{0.6} &
      \textbf{1.7$^{*,\dag}$} &
      10.3 &
      9.8 &
      7.8 &
      8.2 &
      \textbf{7.5} &
      \textbf{8.4$^{*,\dag}$} \\
    SR (\%) $\uparrow$ &
      \multicolumn{2}{c}{100.0} &
      \multicolumn{2}{c}{100.0} &
      \multicolumn{2}{c|}{100.0$\quad$} &
      \multicolumn{2}{c}{100.0} &
      \multicolumn{2}{c}{100.0} &
      \multicolumn{2}{c}{100.0$\quad$} \\ \hline
    \multicolumn{13}{l}{\begin{tabular}[c]{@{}l@{}} \scriptsize{Symbols $*$ and $\dag$ indicate statistical significance ($p \leq 0.05$) with respect to the MPC and SAC, respectively. For different} \\ \scriptsize{metrics, $\uparrow$ means the higher the better, and $\downarrow$ contrarily. Results in bold denote the best.}\end{tabular}} 
    \end{tabular}%
    }
    \end{table*}

    \begin{table*}[!htbp]
    \centering
    \renewcommand\arraystretch{1.05}
    \caption{Scenario C Metrics in the General Case}
    \label{tab:personalized_c}
    \resizebox{0.97 \textwidth }{!}{%
    \begin{tabular}{lr@{$\pm$}lr@{$\pm$}lr@{$\pm$}l|r@{$\pm$}lr@{$\pm$}lr@{$\pm$}l}
    \hline
    \multicolumn{1}{c}{\multirow{2}{*}{\begin{tabular}[c]{@{}c@{}}Scenario C\\ Metrics\end{tabular}}} &
      \multicolumn{6}{c|}{Announced} &
      \multicolumn{6}{c}{Unannounced} \\ \cline{2-13} 
    \multicolumn{1}{c}{} &
      \multicolumn{2}{c}{MPC} &
      \multicolumn{2}{c}{SAC} &
      \multicolumn{2}{c|}{\textbf{HyCPAP$\quad$}} &
      \multicolumn{2}{c}{MPC} &
      \multicolumn{2}{c}{SAC} &
      \multicolumn{2}{c}{\textbf{HyCPAP$\quad$}} \\ \hline
    TBR 2 (\%) $\downarrow$ &
      0.0 &
      0.0 &
      0.0 &
      0.0 &
      0.0 &
      0.0 &
      0.0 &
      0.0 &
      0.0 &
      0.0 &
      0.0 &
      0.0 \\
    TBR 1 (\%) $\downarrow$ &
      0.0 &
      0.0 &
      0.0 &
      0.2 &
      0.0 &
      0.0 &
      0.0 &
      0.0 &
      0.0 &
      0.0 &
      0.0 &
      0.0 \\
    TIR (\%) $\uparrow$ &
      92.4 &
      9.7$\ \ $ &
      96.2 &
      4.3$^*$$\ $ &
      \textbf{97.3} &
      \textbf{3.5$^{*,\dag}$$\ $} &
      77.0 &
      13.3 &
      \textbf{82.8} &
      \textbf{10.2$^*$} &
      \textbf{82.9} &
      \textbf{14.5$^*$} \\
    TAR 1 (\%) $\downarrow$ &
      7.6 &
      9.7 &
      3.8 &
      4.3$^*$ &
      \textbf{2.7} &
      \textbf{3.5$^{*,\dag}$} &
      23.0 &
      13.3 &
      \textbf{17.2} &
      \textbf{10.2$^*$} &
      \textbf{17.1} &
      \textbf{14.5$^*$} \\
    TAR 2 (\%) $\downarrow$ &
      0.1 &
      0.4 &
      \textbf{0.0} &
      \textbf{0.0$^*$} &
      \textbf{0.0} &
      \textbf{0.0$^*$} &
      0.8 &
      1.7 &
      0.3 &
      0.7$^*$ &
      \textbf{0.0} &
      \textbf{0.2$^{*,\dag}$} \\
    SR (\%) $\uparrow$ &
      \multicolumn{2}{c}{100.0} &
      \multicolumn{2}{c}{100.0} &
      \multicolumn{2}{c|}{100.0$\quad$} &
      \multicolumn{2}{c}{100.0} &
      \multicolumn{2}{c}{100.0} &
      \multicolumn{2}{c}{100.0$\quad$} \\ \hline
      \multicolumn{13}{l}{\begin{tabular}[c]{@{}l@{}} \scriptsize{Symbols $*$ and $\dag$ indicate statistical significance ($p \leq 0.05$) with respect to the MPC and SAC, respectively. For different} \\ \scriptsize{metrics, $\uparrow$ means the higher the better, and $\downarrow$ contrarily. Results in bold denote the best.}\end{tabular}} 
    \end{tabular}%
    }
    \end{table*}

    \begin{figure*}[htbp]
    \centering
    \subfloat[]{\label{fig:individual_a}\includegraphics[width=0.3 \textwidth]
    {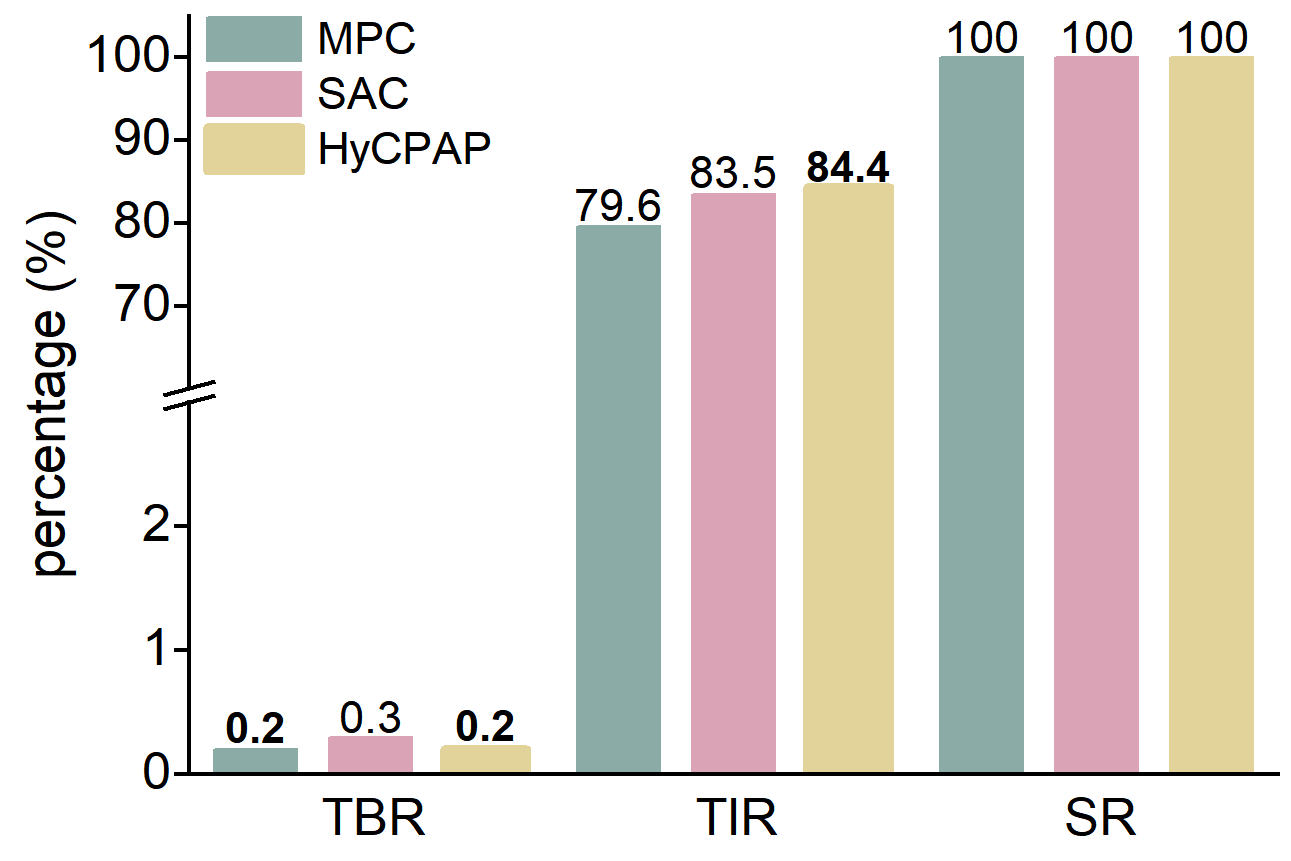}}\hfill
    \subfloat[]{\label{fig:individual_b}\includegraphics[width=0.233 \textwidth, height=0.23\textwidth]{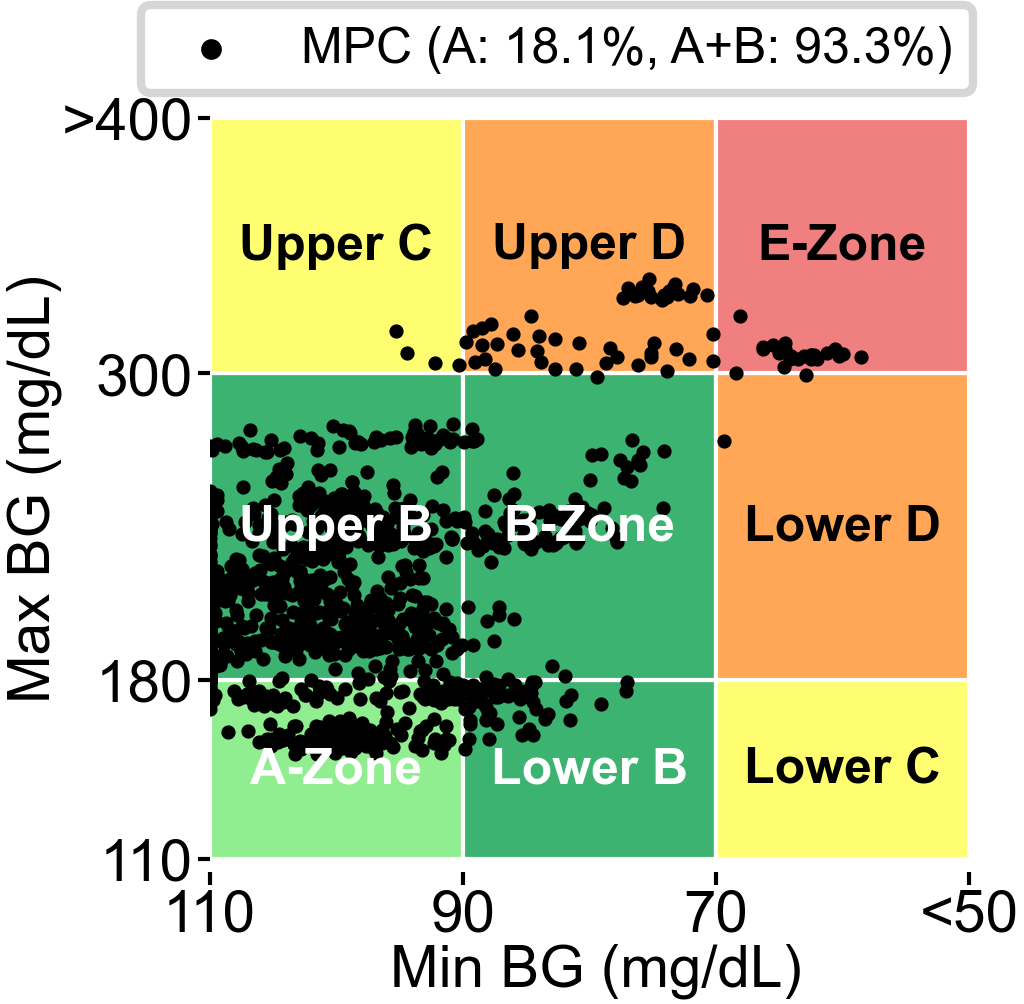}}\hfill
    \subfloat[]{\label{fig:individual_c}\includegraphics[width=0.233 \textwidth, height=0.23\textwidth]{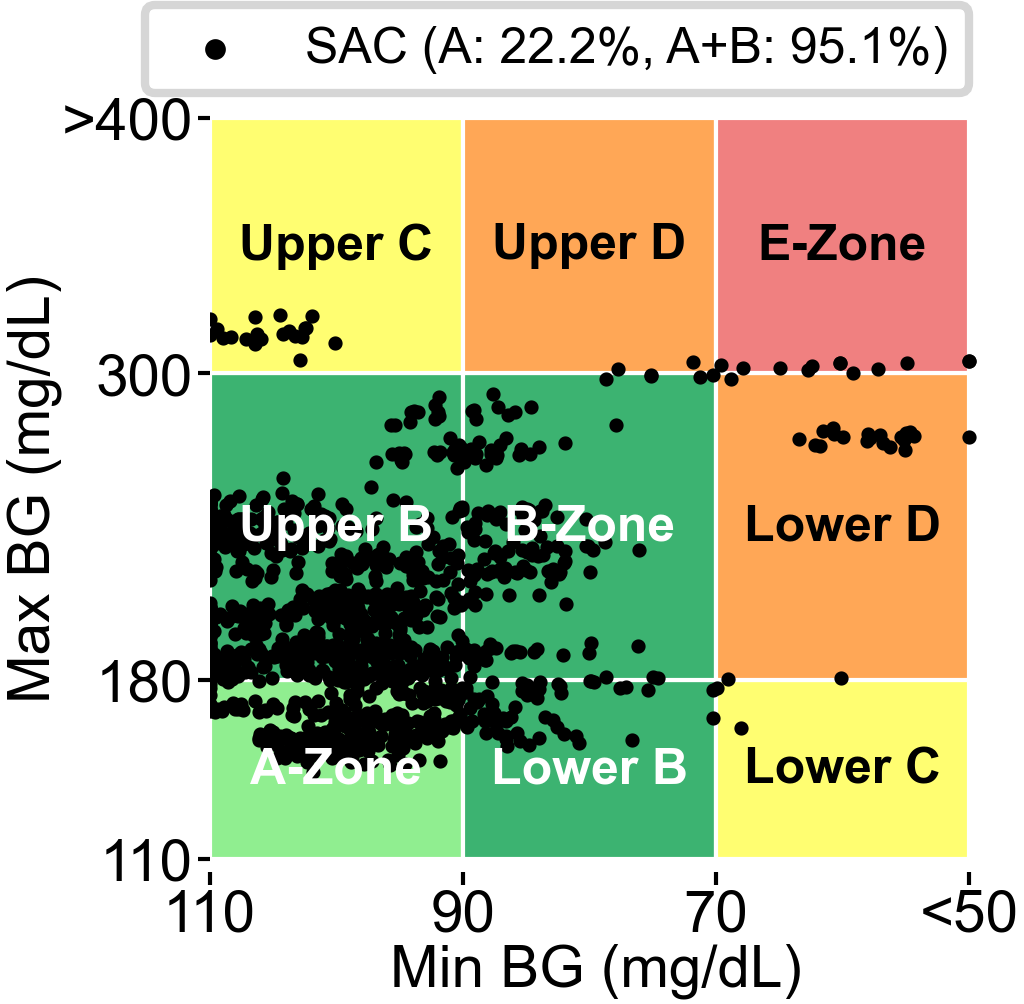}}\hfill
    \subfloat[]{\label{fig:individual_d}\includegraphics[width=0.233 \textwidth, height=0.23\textwidth]{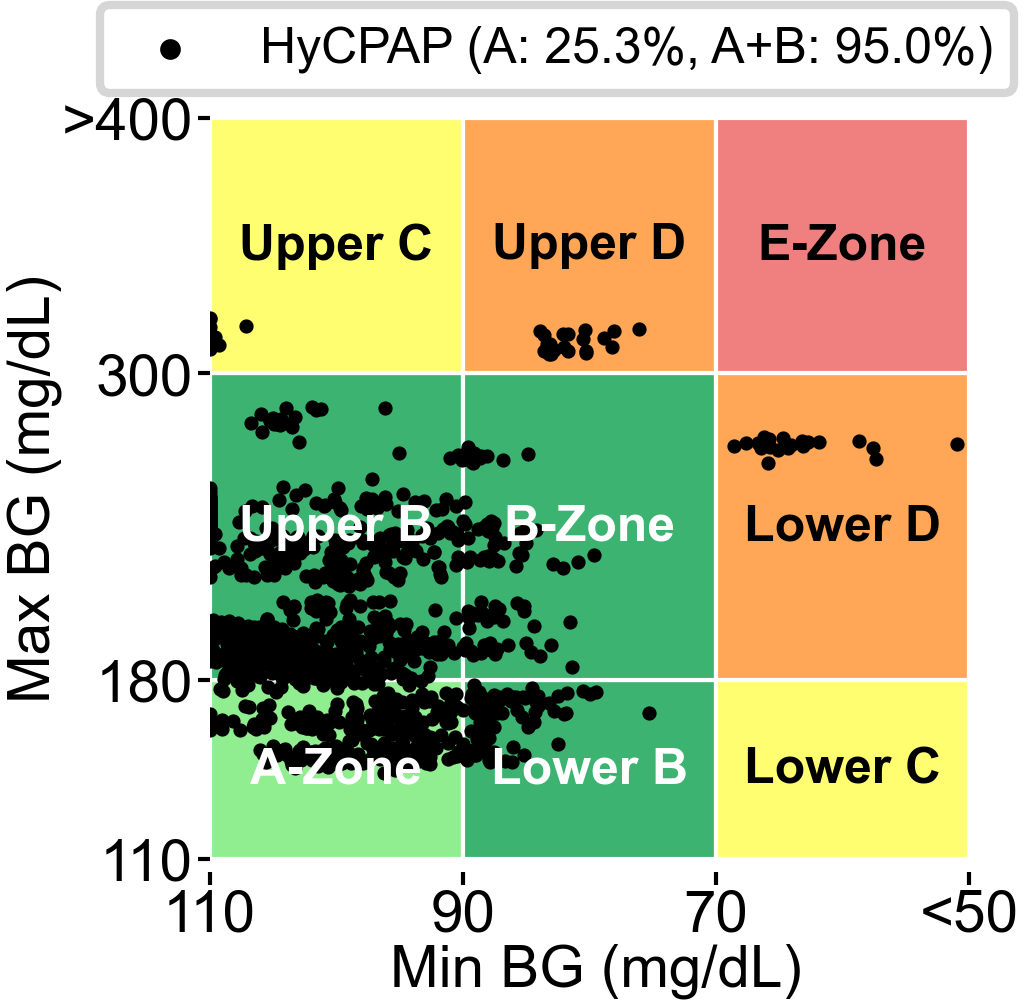}}
    \caption{Overall comparison of methods in the general case. (a): Time below range (TBR), time in range (TIR), and success rate (SR) of the three methods: MPC, SAC, and our proposed HyCPAP. The best-performing results are highlighted in bold. (b)-(d): The CVGA plot of MPC, SAC, and our proposed HyCPAP, respectively.}
    \label{fig:individual}
    \end{figure*}      

    \begin{table*}[!htbp]
    \centering
    \renewcommand\arraystretch{1.1}
    \caption{Scenario A Metrics in the Data-limited Case}
    \label{tab:meta_a}
    \resizebox{0.96 \textwidth}{!}{%
    \begin{tabular}{lr@{$\pm$}lr@{$\pm$}lr@{$\pm$}l|r@{$\pm$}lr@{$\pm$}lr@{$\pm$}l}
    \hline
    \multicolumn{1}{c}{\multirow{2}{*}{\begin{tabular}[c]{@{}c@{}}Scenario A\\ Metrics\end{tabular}}} &
      \multicolumn{6}{c|}{Announced} &
      \multicolumn{6}{c}{Unannounced} \\ \cline{2-13} 
    \multicolumn{1}{c}{} &
      \multicolumn{2}{c}{MPC} &
      \multicolumn{2}{c}{\textbf{Meta-SAC}} &
      \multicolumn{2}{c|}{\textbf{Meta-HyCPAP}} &
      \multicolumn{2}{c}{MPC} &
      \multicolumn{2}{c}{SAC} &
      \multicolumn{2}{c}{\textbf{Meta-HyCPAP}$\quad$} \\ \hline
    TBR 2 (\%) $\downarrow$ &
      0.0 &
      0.0 &
      0.0 &
      0.0 &
      0.0 &
      0.0 &
      0.0 &
      0.0 &
      0.0 &
      0.0 &
      0.0 &
      0.0 \\
    TBR 1 (\%) $\downarrow$ &
      0.0 &
      0.0 &
      0.0 &
      0.0 &
      0.0 &
      0.0 &
      0.0 &
      0.0 &
      0.0 &
      0.0 &
      0.0 &
      0.0 \\
    TIR (\%) $\uparrow$ &
      92.6 &
      9.4$\ $&
      \textbf{94.4} &
      \textbf{7.5$^*$} &
      \textbf{94.0} &
      \textbf{8.8$^* $} &
      77.4 &
      13.2 &
      \textbf{80.0} &
      \textbf{12.9$^*$} &
      \textbf{81.4} &
      \textbf{14.9$^{*,\dag}$} \\
    TAR 1 (\%) $\downarrow$ &
      7.4 &
      9.4 $\ $&
      \textbf{5.6} &
      \textbf{7.5$^*$$\ $} &
      \textbf{6.0} &
      \textbf{8.8$^*$$\ $} &
      22.6 &
      13.2 &
      \textbf{20.0} &
      \textbf{12.9$^*$} &
      \textbf{18.6} &
      \textbf{14.9$^{*,\dag}$} \\
    TAR 2 (\%) $\downarrow$ &
      0.0 &
      0.0 &
      0.0 &
      0.0 &
      0.0 &
      0.0 &
      0.4 &
      0.9 &
      \textbf{0.2} &
      \textbf{0.6$^*$} &
      \textbf{0.2} &
      \textbf{0.7$^*$} \\
    SR (\%) $\uparrow$ &
      \multicolumn{2}{c}{100.0} &
      \multicolumn{2}{c}{100.0} &
      \multicolumn{2}{c|}{100.0$\quad\ $} &
      \multicolumn{2}{c}{100.0} &
      \multicolumn{2}{c}{100.0} &
      \multicolumn{2}{c}{100.0$\quad\quad\ $} \\ \hline
      \multicolumn{13}{l}{\begin{tabular}[c]{@{}l@{}} \scriptsize{Symbols $*$ and $\dag$ indicate statistical significance ($p \leq 0.05$) with respect to the MPC and Meta-SAC, respectively. For different } \\ \scriptsize{metrics, $\uparrow$ means the higher the better, and $\downarrow$ contrarily. Results in bold denote the best.} \end{tabular}} 
    \end{tabular}%
    }
    \end{table*}

    \begin{table*}[!htbp]
        \centering
        \renewcommand\arraystretch{1.1}
        \caption{Scenario B Metrics in the Data-limited Case}
        \label{tab:meta_b}
        \resizebox{0.97 \textwidth}{!}{%
        \begin{tabular}{lr@{$\pm$}lr@{$\pm$}lr@{$\pm$}l|r@{$\pm$}lr@{$\pm$}lr@{$\pm$}l}
        \hline
        \multicolumn{1}{c}{\multirow{2}{*}{\begin{tabular}[c]{@{}c@{}}Scenario B\\ Metrics\end{tabular}}} &
          \multicolumn{6}{c|}{Announced} &
          \multicolumn{6}{c}{Unannounced} \\ \cline{2-13} 
        \multicolumn{1}{c}{} &
          \multicolumn{2}{c}{MPC} &
          \multicolumn{2}{c}{Meta-SAC} &
          \multicolumn{2}{c|}{\textbf{Meta-HyCPAP}} &
          \multicolumn{2}{c}{MPC} &
          \multicolumn{2}{c}{Meta-SAC} &
          \multicolumn{2}{c}{\textbf{Meta-HyCPAP$\quad$}} \\ \hline
        TBR 2 (\%) $\downarrow$ &
          \textbf{0.0} &
          \textbf{0.0} &
          0.8 &
          0.8$^*$ &
          \textbf{0.0} &
          \textbf{0.0$^{\dag}$} &
          \textbf{0.0} &
          \textbf{0.0} &
          9.0 &
          9.0$^*$ &
          \textbf{0.0} &
          \textbf{0.0$^{\dag}$} \\
        TBR 1 (\%) $\downarrow$ &
          \textbf{1.1} &
          \textbf{3.4} &
          1.9 &
          5.4$^*$ &
          \textbf{1.2} &
          \textbf{3.6$^{\dag}$} &
          \textbf{0.1} &
          \textbf{0.6} &
          10.3 &
          23.4$^*$ &
          \textbf{0.1} &
          \textbf{0.6$^{\dag}$} \\
        TIR (\%) $\uparrow$ &
          80.5 &
          19.8 &
          80.5 &
          21.6 &
          80.8 &
          20.1 &
          57.8 &
          11.8 &
          51.5 &
          19.9$^*$ &
          \textbf{59.1} &
          \textbf{12.8$^{*,\dag}$} \\
        TAR1 (\%) $\downarrow$ &
          18.4 &
          17.5 &
          \textbf{17.5} &
          \textbf{17.6$^*$} &
          18.0 &
          17.7$^{*,\dag}$ &
          42.1 &
          11.8 &
          \textbf{38.2} &
          \textbf{16.3$^*$} &
          40.8 &
          12.7$^{*,\dag}$ \\
        TAR2 (\%) $\downarrow$ &
          2.0 &
          5.3 &
          \textbf{1.5} &
          \textbf{4.3$^*$} &
          \textbf{1.6} &
          \textbf{4.7$^{*}$} &
          10.3 &
          9.8 &
          9.8 &
          10.8 &
          10.5 &
          10.0 \\
        SR (\%) $\uparrow$ &
          \multicolumn{2}{c}{\textbf{100.0}} &
          \multicolumn{2}{c}{99.5} &
          \multicolumn{2}{c|}{\textbf{100.0$\quad\ $}} &
          \multicolumn{2}{c}{\textbf{100.0}} &
          \multicolumn{2}{c}{80.0} &
          \multicolumn{2}{c}{\textbf{100.0$\quad\quad\ $}} \\ \hline
          \multicolumn{13}{l}{\begin{tabular}[c]{@{}l@{}} \scriptsize{Symbols $*$ and $\dag$ indicate statistical significance ($p \leq 0.05$) with respect to the MPC and Meta-SAC, respectively. For different } \\ \scriptsize{metrics, $\uparrow$ means the higher the better, and $\downarrow$ contrarily. Results in bold denote the best.} \end{tabular}} 
        \end{tabular}%
        }
        \end{table*}

    \begin{table*}[]
    \centering
    \renewcommand\arraystretch{1.1}
    \caption{Scenario C Metrics in the Data-limited Case}
    \label{tab:meta_c}
    \resizebox{0.97 \textwidth}{!}{%
    \begin{tabular}{lr@{$\pm$}lr@{$\pm$}lr@{$\pm$}l|r@{$\pm$}lr@{$\pm$}lr@{$\pm$}l}
    \hline
    \multicolumn{1}{c}{\multirow{2}{*}{\begin{tabular}[c]{@{}c@{}}Scenario C\\ Metrics\end{tabular}}} &
      \multicolumn{6}{c|}{Announced} &
      \multicolumn{6}{c}{Unannounced} \\ \cline{2-13} 
    \multicolumn{1}{c}{} &
      \multicolumn{2}{c}{MPC} &
      \multicolumn{2}{c}{Meta-SAC} &
      \multicolumn{2}{c|}{\textbf{Meta-HyCPAP}} &
      \multicolumn{2}{c}{MPC} &
      \multicolumn{2}{c}{\textbf{Meta-SAC}} &
      \multicolumn{2}{c}{\textbf{Meta-HyCPAP$\quad$}} \\ \hline
    TBR 2 (\%) $\downarrow$ &
      0.0 &
      0.0 &
      0.0 &
      0.0 &
      0.0 &
      0.0 &
      0.0 &
      0.0 &
      0.0 &
      0.0 &
      0.0 &
      0.0 \\
    TBR 1 (\%) $\downarrow$ &
      0.0 &
      0.0 &
      0.0 &
      0.0 &
      0.0 &
      0.1 &
      0.0 &
      0.0 &
      0.0 &
      0.0 &
      0.0 &
      0.0 \\
    TIR (\%) $\uparrow$ &
      92.4 &
      9.7$\ $ &
      92.9 &
      9.5$^*$$\ $&
      \textbf{93.6} &
      \textbf{9.3$^{*,\dag}$ $\ $} &
      77.0 &
      13.3 &
      \textbf{79.8} &
      \textbf{12.3$^*$} &
      \textbf{80.1} &
      \textbf{15.3$^*$} \\
    TAR1 (\%) $\downarrow$ &
      7.6 &
      9.7 &
      7.1 &
      9.5$^*$ &
      \textbf{6.4} &
      \textbf{9.3$^{*,\dag}$} &
      23.0 &
      13.3 &
      \textbf{20.2} &
      \textbf{12.3$^*$} &
      \textbf{19.9} &
      \textbf{15.3$^*$} \\
    TAR2 (\%) $\downarrow$ &
      0.1 &
      0.4 &
      0.0 &
      0.0 &
      0.0 &
      0.1 &
      0.8 &
      1.7 &
      \textbf{0.3} &
      \textbf{0.6$^*$} &
      \textbf{0.3} &
      \textbf{0.7$^*$} \\
    SR (\%) $\uparrow$ &
      \multicolumn{2}{c}{100.0} &
      \multicolumn{2}{c}{100.0} &
      \multicolumn{2}{c|}{100.0$\quad\ $} &
      \multicolumn{2}{c}{100.0} &
      \multicolumn{2}{c}{100.0} &
      \multicolumn{2}{c}{100.0$\quad\quad\ $} \\ \hline
      \multicolumn{13}{l}{\begin{tabular}[c]{@{}l@{}} \scriptsize{Symbols $*$ and $\dag$ indicate statistical significance ($p \leq 0.05$) with respect to the MPC and Meta-SAC, respectively. For different } \\ \scriptsize{metrics, $\uparrow$ means the higher the better, and $\downarrow$ contrarily. Results in bold denote the best.} \end{tabular}} 
        \end{tabular}%
        }
        \end{table*}
        \begin{figure*}[!htbp]
        \centering
        \subfloat[]{\label{fig:meta_a}\includegraphics[width=0.3 \textwidth ]{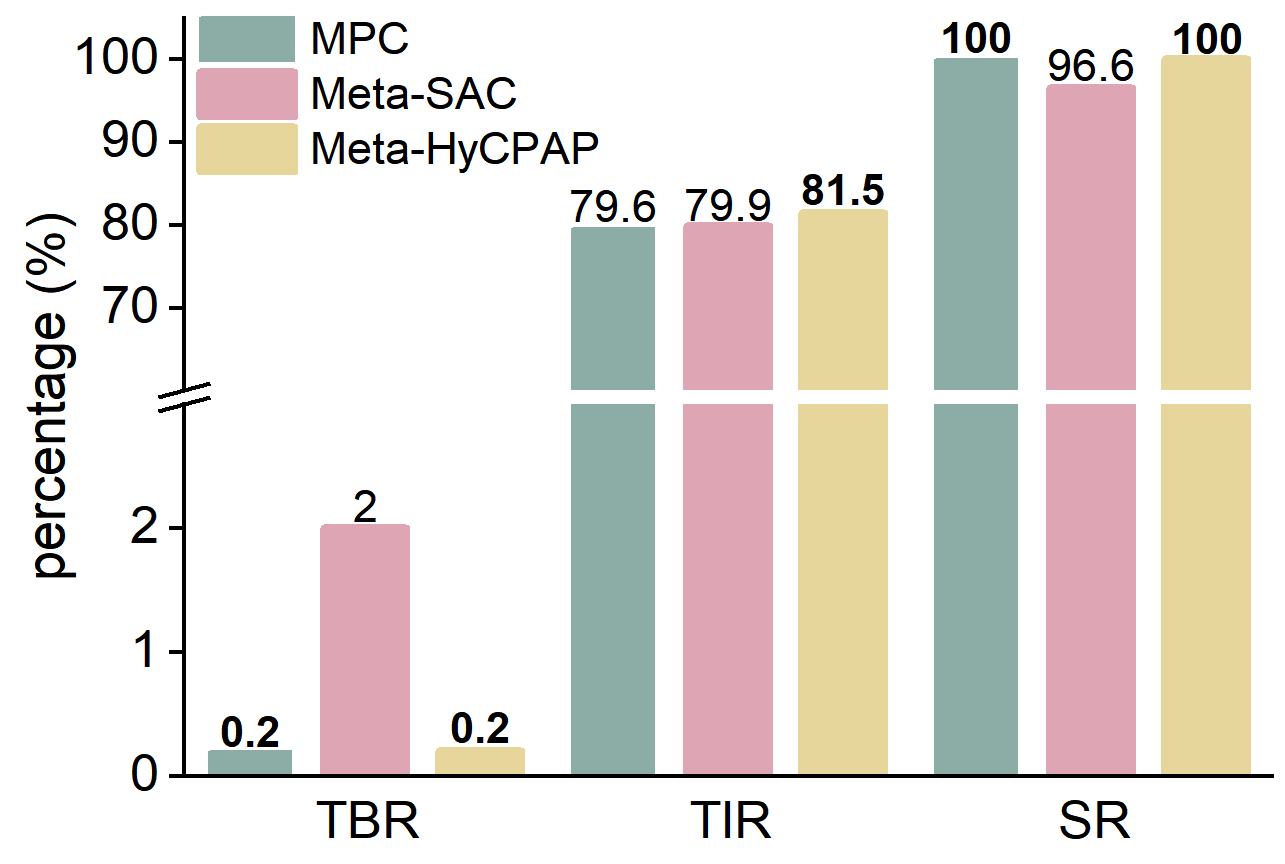}}\hfill
        \subfloat[]{\label{fig:meta_b}\includegraphics[width=0.233 \textwidth, height=0.23\textwidth]{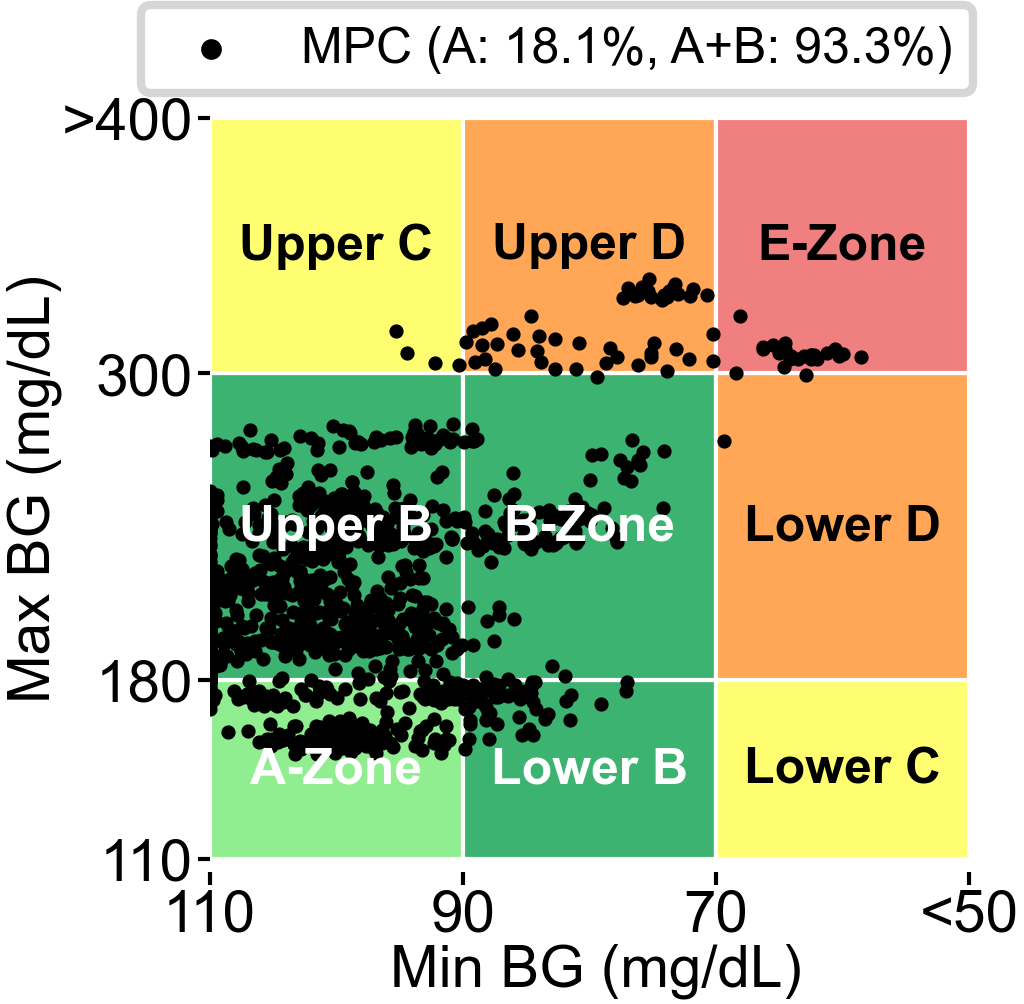}}\hfill
        \subfloat[]{\label{fig:meta_c}\includegraphics[width=0.233 \textwidth, height=0.23\textwidth]{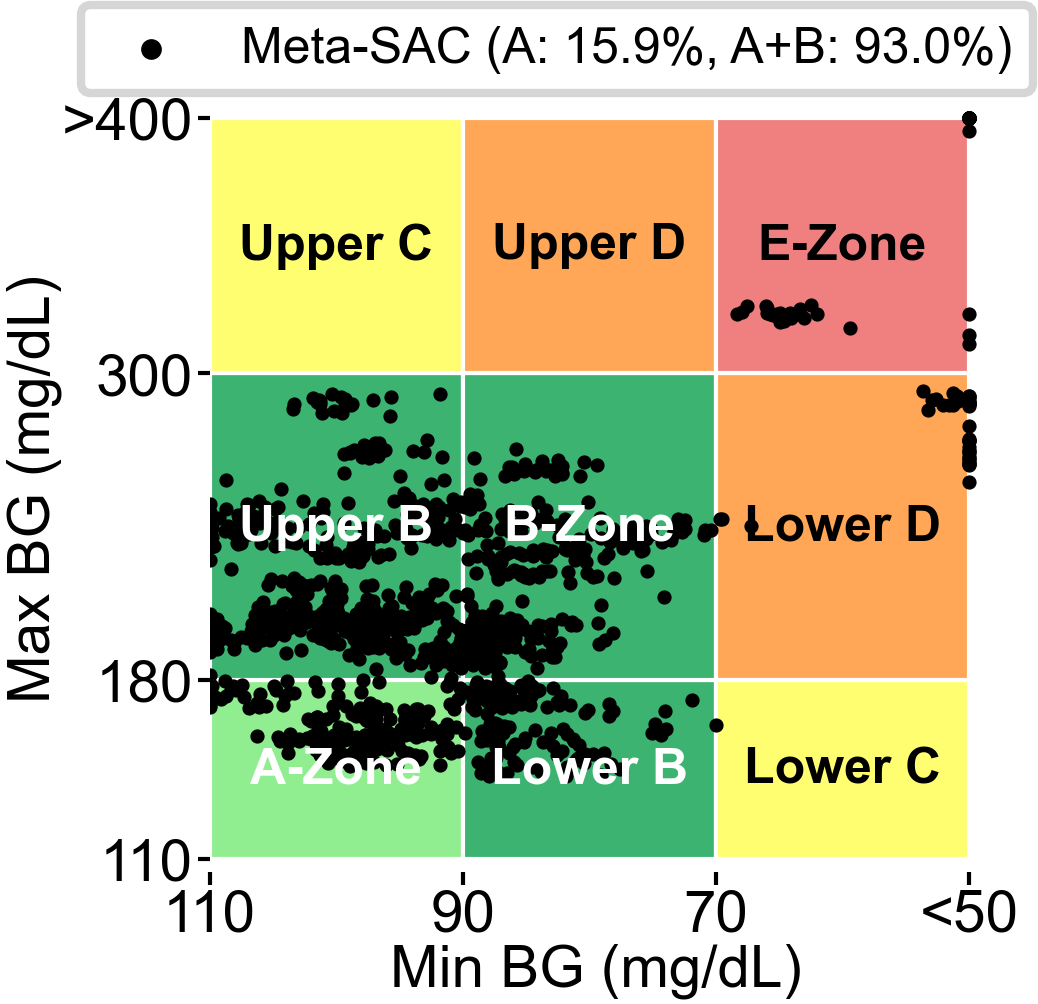}}\hfill
        \subfloat[]{\label{fig:meta_d}\includegraphics[width=0.233 \textwidth, height=0.23\textwidth]{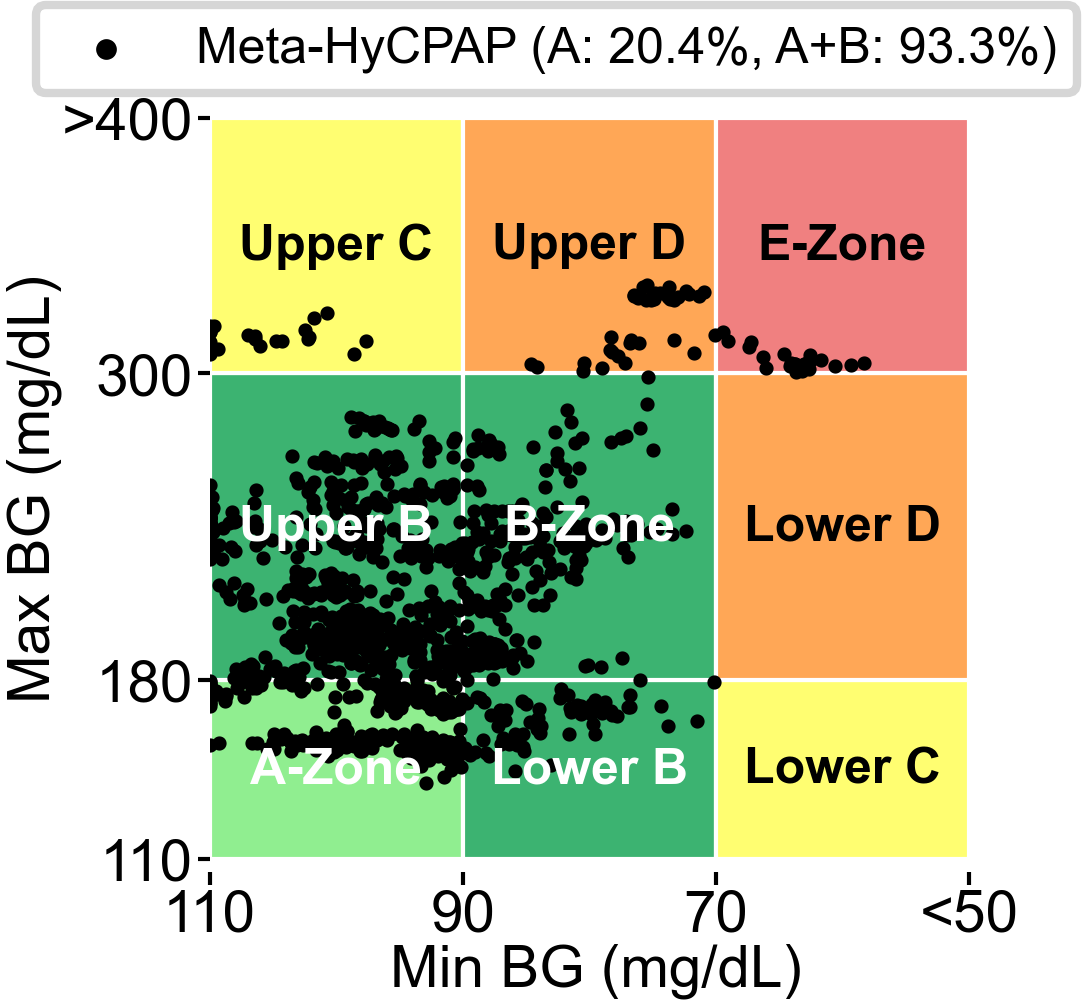}}
        \caption{Overall comparison of methods in the data-limited case. (a): Time below range (TBR), time in range (TIR), and success rate (SR) of the three methods: MPC, Meta-SAC, and our proposed Meta-HyCPAP. The best-performing results are highlighted in bold. (b)-(d): The CVGA plot of MPC, Meta-SAC, and our proposed Meta-HyCPAP, respectively.}
        \label{fig:meta}
        \end{figure*}

            Scenario B evaluates the stability of the control algorithms in protocol with extremely large meal sizes that DRL policies have never encountered during training. From Table \ref{tab:personalized_b}, we observe that the SAC algorithm, owing to the presence of the distribution shift problem, leads to the highest TBR in both situations. And the BG level below the desired range can be potentially fatal to patients. However, HyCPAP effectively addresses the distribution shift problem by behaving similarly to the MPC policy when the DRL policy lacks confidence. It ensures secure and reliable performance in the unfamiliar Scenario B. In addition to reducing hypoglycemia, HyCPAP also benefits from the flexibility and adaptability of the DRL policy, resulting in the highest TIR while achieving the lowest TAR 1 and TAR 2, in both announced and unannounced meal situations.
            
            Scenario C evaluates the performance of control algorithms to handle meal uncertainty, encompassing variations in meal timing and size. The corresponding results are presented in Table \ref{tab:personalized_c}. In situations where meals are announced, our HyCPAP method outperforms both the MPC and SAC algorithms, achieving the highest TIR and the lowest TAR values. When faced with unannounced meals, the proposed HyCPAP performs comparably to SAC, but effectively mitigating severe hyperglycemia. Notably, both HyCPAP and SAC outperform the MPC policy in this context. These results demonstrate the efficacy of HyCPAP to enhance glucose control in uncertain conditions.
            
        \subsubsection{Overall Comparison}
            The overall comparison in the general case is shown in Fig. \ref{fig:individual}. Our proposed HyCPAP outperforms the MPC algorithms in terms of the TIR (\textbf{84.4\%} vs. 79.6\% and 83.5\%) and achieves the same TBR with MPC (\textbf{0.2\%}) which is better than SAC (0.4\%) as demonstrated in Fig. \ref{fig:individual_a}. Additionally, Fig. \ref{fig:individual_b}, \ref{fig:individual_c}, and \ref{fig:individual_d} further validate the superiority of the proposed HyCPAP method. Specifically, our method and SAC achieve similar percentages in the A+B zone but higher than MPC. Significantly, our method achieves the highest percentage in the A zone (\textbf{25.3\%} vs. 22.2\% and 18.1\%). The obtained results strongly support the superiority of HyCPAP in achieving effective glucose management through the combination of MPC and DRL policies, offering personalized and adaptive capabilities while ensuring safety and stability.
            
    \subsection{Performance in the Data-limited Case}
    In scenarios where data availability is limited, we aim to enable fast adaptation of the control policy to new patients using only a small amount of data. The results demonstrate that our proposed approach, Meta-HyCPAP, effectively leverages previous experience and shared knowledge derived from known patients. This allows for additional training on the learned policy, ultimately enhancing its capacity to regulate glucose levels in new patients.
        \subsubsection{Scenario Comparison}
            Table \ref{tab:meta_a} presents the performance comparison of the proposed Meta-HyCPAP and other baselines methods in scenario A. Meta-HyCPAP and Meta-SAC policy show similar performance without any significance but they both outperform the MPC policy with announced meals. However, in situations where meals are not announced, Meta-HyCPAP achieves the highest TIR, and lowest TAR 1. This shows that Meta-HyCPAP effectively increases TIR for new patients, without raising the risk of hyperglycemia.
            
            From Table \ref{tab:meta_b}, we observe that the Meta-SAC algorithm leads to a higher occurrence of hypoglycemia and severe hypoglycemia, occasionally even resulting in critical failure. 
            In contrast, Meta-HyCPAP comparably prevents hypoglycemia and ensures patient safety by behaving similarly to MPC policy in unseen states. And it leverages RL and meta-learning for personalized insulin delivery with limited data, resulting in higher TIR without increased hypoglycemia and hyperglycemia risk, especially during unannounced meals.
            
            The results in scenario C are shown in Table \ref{tab:meta_c}. In situations where meals are announced, Meta-HyCPAP demonstrates superior performance compared to both the MPC and Meta-SAC algorithms, as evidenced by its higher TIR and lower TAR 1. In unannounced meal situations, the Meta-HyCPAP method performs comparably to Meta-SAC, with both algorithms surpassing the performance of MPC. These results highlight the effectiveness of Meta-HyCPAP in handling uncertainty and underscore its potential for enhancing glycemic control in varying conditions.

        \subsubsection{Overall Comparison}
            In the data-limited case, Fig. \ref{fig:meta} demonstrates the superiority of our Meta-HyCPAP. It outperforms other algorithms in TIR (\textbf{81.5\%} vs. 79.6\% and 79.9\%) and achieves the same TBR as MPC (\textbf{0.2\%}), surpassing Meta-SAC (2.0\%) as shown in Fig. \ref{fig:meta_a}. Additionally, Fig. \ref{fig:meta_b}, \ref{fig:meta_c}, and \ref{fig:meta_d} further validate the superiority of our hybrid method. Notably, our Meta-HyCPAP achieves the highest percentage in the A zone (\textbf{20.4\%} vs. 18.1\% and 15.9\%), while both our method and MPC perform well in the A+B zone, surpassing Meta-SAC. These results demonstrate the effective integration of meta-learning techniques in Meta-HyCPAP, enabling fast adaptation of the policy to new patients with limited data.
            
\section{Conclusion}
\label{sec:conclusion}
In this paper, we present HyCPAP, a novel hybrid control policy for the AP system, to leverage the strengths of MPC and DRL policies and compensate for their individual shortcomings. By merging MPC policy with an ensemble DRL policy, HyCPAP offers personalized and adaptive capabilities while ensuring safety and stability. Additionally, to facilitate faster deployment of AP systems in real-world settings, we further incorporate meta-learning techniques into HyCPAP, leveraging previous experience and patient-shared knowledge. This allows our policy to fast adapt to new patients using limited available data. Through extensive evaluations using the FDA-accepted UVA/Padova T1DM simulator \cite{Simulator}, our proposed methods demonstrate their effectiveness in glucose management, achieving the highest percentage of time spent in the desired euglycemic range, and lowest occurrences of hypoglycemia in both general and data-limited case. These results indicate the great potential of our methods for closed-loop glucose control in individuals with T1DM.

Although our approaches achieve superior control performance in \textit{silico}. However, physical exercise is not considered in this study, posing a challenge for a fully automated AP system. Another limitation is the independent operation of MPC and DRL policies without mutual information exchange in the fusion procedure. Future research will focus on considering physical activity and enhancing the hybrid policy through knowledge sharing.

\bibliographystyle{IEEEtran}
\bibliography{generic-color.bib}

\end{document}